\documentclass[pdflatex,sn-mathphys-num]{sn-jnl}

\usepackage{graphicx}
\usepackage{multirow}
\usepackage{amsmath,amssymb,amsfonts}
\usepackage{amsthm}
\usepackage{mathrsfs}
\usepackage[title]{appendix}
\usepackage{xcolor}
\usepackage{textcomp}
\usepackage{manyfoot}
\usepackage{booktabs}
\usepackage{algorithm}
\usepackage{algorithmicx}
\usepackage{algpseudocode}
\usepackage{listings}
\usepackage{subcaption}
\usepackage{makecell}

\theoremstyle{thmstyleone}

\theoremstyle{thmstyletwo}

\theoremstyle{thmstylethree}

\definecolor{apricot}{RGB}{0,204,0}
\graphicspath{{../wacv_2026_final/images/}}

\begin{document}

\title[Vision LLMs for Engagement Analysis]{Vision Large Language Models Are Good Noise Handlers in Engagement Analysis}

\author[1]{\fnm{Alexander} \sur{Vedernikov}}\email{aleksandr.vedernikov@oulu.fi}
\author[1]{\fnm{Puneet} \sur{Kumar}}\email{puneet.kumar@oulu.fi}
\author[1]{\fnm{Haoyu} \sur{Chen}}\email{chen.haoyu@oulu.fi}
\author[1]{\fnm{Tapio} \sur{Seppänen}}\email{tapio.seppanen@oulu.fi}
\author*[1,2]{\fnm{Xiaobai} \sur{Li}}\email{xiaobai.li@zju.edu.cn}

\affil*[1]{\orgdiv{Center for Machine Vision and Signal Analysis}, \orgname{University of Oulu}, \orgaddress{\city{Oulu}, \country{Finland}}}
\affil[2]{\orgdiv{State Key Laboratory of Blockchain and Data Security}, \orgname{Zhejiang University}, \orgaddress{\city{Hangzhou}, \country{China}}}

\abstract{Engagement recognition in video datasets, unlike traditional image classification tasks, is particularly challenged by subjective labels and noise limiting model performance. To overcome the challenges of subjective and noisy engagement labels, we propose a framework leveraging Vision Large Language Models (VLMs) to refine annotations and guide the training process. Our framework uses a questionnaire to extract behavioral cues and split data into high- and low-reliability subsets. We also introduce a training strategy combining curriculum learning with soft label refinement, gradually incorporating ambiguous samples while adjusting supervision to reflect uncertainty. We demonstrate that classical computer vision models trained on refined high-reliability subsets and enhanced with our curriculum strategy show improvements, highlighting benefits of addressing label subjectivity with VLMs. This method surpasses prior state of the art across engagement benchmarks such as \textit{EngageNet} (three of six feature settings, maximum improvement of $+1.21\%$), and \textit{DREAMS} / \textit{PAFE} with F1 gains of $+0.22$ / $+0.06$.}


\maketitle

\section{Introduction}\label{sec:intro}
Affective computing has driven progress in emotion recognition across psychology, education, and human--computer interaction \cite{Noroozi2021505}. While Ekman’s basic emotions \cite{ekman1992argument} are well studied, non-typical emotions like engagement remain underexplored \cite{kumar2024nontypical}. In this context, engagement recognition refers to estimating a learner’s level of involvement from behavioral cues such as gaze, posture, facial affect, and distractions. Video-based engagement recognition seeks to predict this level from short video segments of learners interacting with instructional material. Engagement’s inherent subjectivity and sparse human-generated annotations introduce label disagreements, leading to noise \cite{plank2022problemhumanlabelvariation,Kumar_2025_WACV}, overfitting, reduced generalization \cite{Tu_2023_CVPR,wen2023benign}. Addressing these challenges and improving dataset quality through annotation refinement \cite{kumar2022vistanet} is vital for effective engagement recognition \cite{Vedernikov_2024_CVPR}.

Approaches to Learning with Noisy Labels (LNL), such as module-based \cite{Chen20222106}, loss-based \cite{khan2023novellossfunctionutilizing}, label correction \cite{Zhou_PR_PL_10160130}, and sample selection \cite{Darshan_9607483} methods, attempt to overcome annotation challenges. Yet, in subjective tasks such as emotion recognition \cite{wu2022novel,chou2019every,wu2024can} and engagement recognition, they struggle with annotation variability and data loss \cite{Karim_2022_CVPR}.

\begin{figure}[!h]
    \centering
    \includegraphics[width=0.75\linewidth]{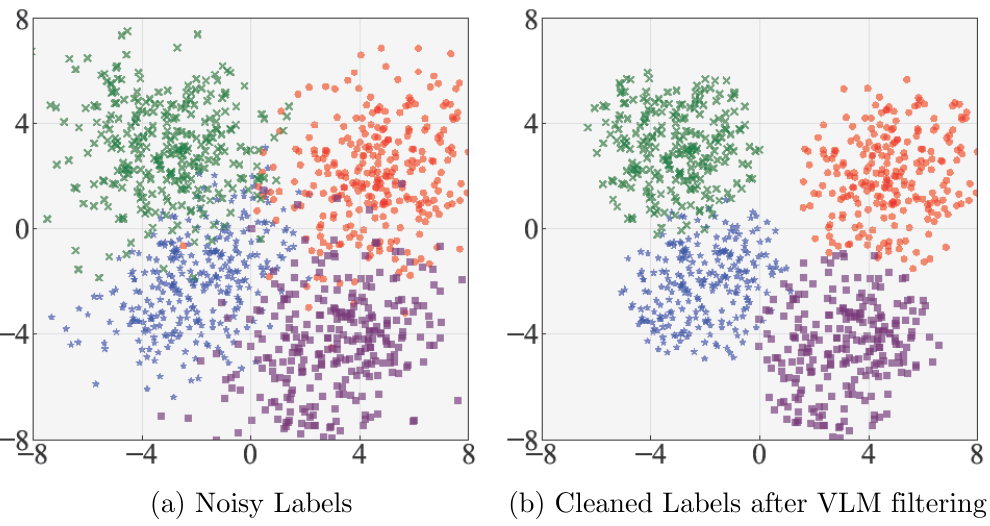}
    \caption{Dataset refinement through VLM analysis. Visualization of noisy vs. cleaned labels, illustrating VLM filtering on dataset quality. Markers represent dataset's different classes. In this scheme, “different classes” correspond to the four engagement levels used throughout our experiments ($0$: Not Engaged, $1$: Barely Engaged, $2$: Engaged, $3$: Highly Engaged), and the plot is illustrative only rather than taken from a specific dataset or VLM configuration.}
    \label{fig:intro}
\end{figure}

VLMs integrate computer vision and natural language processing through transformer advances \cite{mckinzie2024mm1}, addressing these challenges. By posing structured questions about observable cues, VLMs can uncover subtle annotation inconsistencies and mitigate subjectivity in emotion recognition \cite{yang2024emoll}. Their capacity for automated label correction enhances dataset quality and model generalization. Recent studies have also begun to explore VLMs for learning engagement and related academic emotion recognition, but typically use VLMs as direct predictors rather than as tools for structuring label reliability and guiding downstream training \cite{Teotia10917661, wang2025usingvisionlanguagemodels}.

\begin{figure*}[t]
    \centering
    \includegraphics[width=1.00\linewidth]{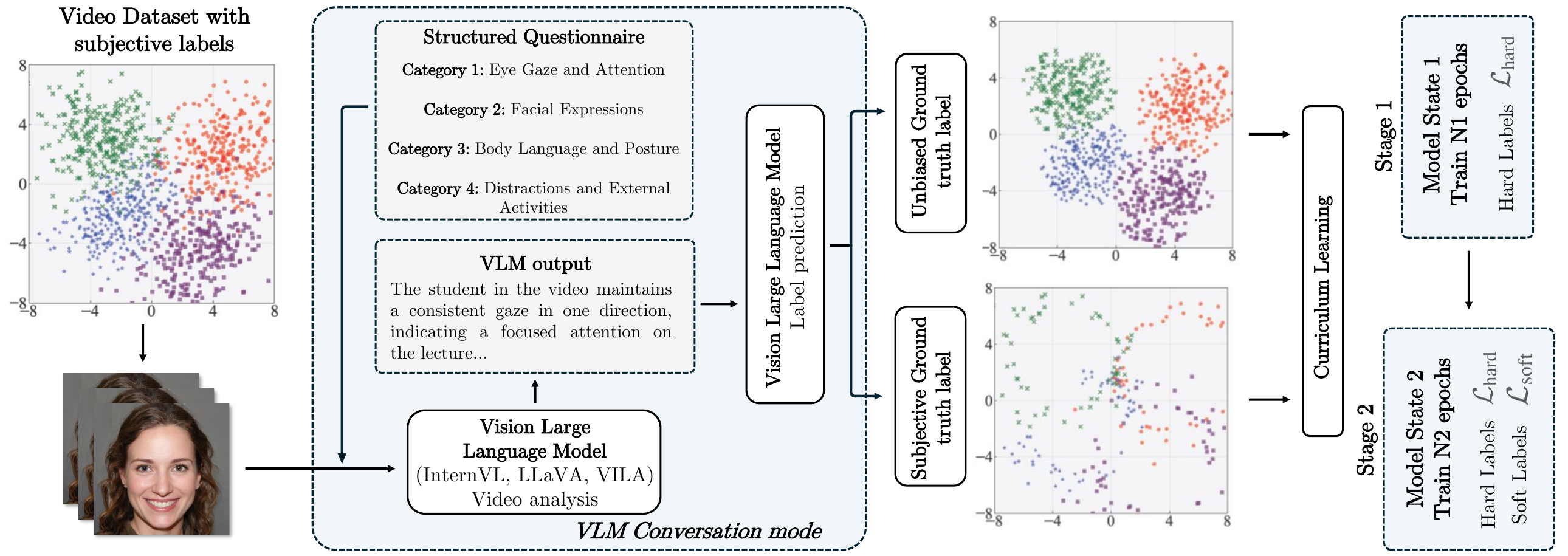}
    \caption{Overview of our method for engagement analysis with subjective labels. A structured questionnaire guides a VLM to assess label reliability, generating informed outputs. It then uses two-stage curriculum learning, first training on high-confidence data and then introducing ambiguous samples with soft labels. This mitigates label noise and enhances model performance.}
    \label{fig:method_main}
\end{figure*}

We present a framework that uses VLMs to improve analysis of engagement datasets (\textit{EngageNet}, \textit{PAFE}, and \textit{DREAMS}) by addressing label subjectivity and noise. We prompt InternVL$2$-$8$B with $15$ structured yes/no questions on $16$ sampled frames per video to extract behavioral cues (eye gaze, facial expression, body posture, and distractions) to detect annotation inconsistencies. Samples where VLM agrees with the original label form a high-reliability (‘Accepted’) subset, while disagreements form a low-reliability (‘Rejected’) subset. We then train classical computer vision models (e.g., TCCT-Net, ResNet+TCN, EfficientNet+LSTM) using a two-stage curriculum: Stage~1 trains on Accepted samples only; Stage~2 adds Rejected samples and implements curriculum learning with soft label refinement during training, allowing the model to gradually learn from less ambiguous samples while substituting hard one-hot labels with uncertainty-aware soft targets for ambiguous samples. During data augmentation, segment sampling is weighted by VLM-derived reliability. Finally, we evaluate accuracy and F$1$ on all three datasets and compare against baseline models. Result of dataset refinement is depicted in Figure~\ref{fig:intro}, while detailed refinement process is explained in Figure~\ref{fig:method_main}. 
The Figure~\ref{fig:intro} is intended as an illustrative schematic of the refinement concept rather than a direct visualization from a particular dataset or VLM run. Concrete datasets (\textit{EngageNet}, \textit{PAFE}, \textit{DREAMS}) and VLM choices are described in detail in Sections~\ref{sec:3_method} and \ref{sec:4_experiments}. Findings indicate that traditional computer vision models, when trained on selected high-reliability subsets and further enhanced by curriculum strategy, achieve noticeable improvements. The contributions are:

\begin{enumerate}
    \item We use VLMs to detect and correct label noise in engagement datasets, thereby enhancing annotation consistency without relying on pre-existing noise models.
    \item By using a structured questionnaire, we guide VLMs to focus on observable engagement cues, thus reducing ambiguity and improving zero- and few-shot performance.
    \item We use VLM-based filtering to partition the dataset into accepted and rejected subsets, which reduces noise and improves the robustness of classical vision models.
    \item Our training strategy incorporates curriculum learning and soft label refinement to progressively include samples with varying levels of ambiguity, addressing the challenges posed by label subjectivity.
    \item Experiments on multiple datasets demonstrate our approach improves accuracy and robustness while matching human tolerance in engagement annotation.
\end{enumerate}

\section{Related Work}\label{sec:2_related_work}
\subsection{Subjectivity in Engagement Recognition}
In contrast to typical emotions with clear indicators, engagement, being non-typical, is subject to ambiguous cues, making its recognition inherently subjective. This subjectivity results in annotations that introduce label noise, weakening model reliability. In \textit{DAiSEE} and similar datasets, inconsistent labeling worsens annotation noise and impairs model generalization \cite{Khan2024111}. Moreover, as engagement is inherently nuanced and multidimensional, traditional binary and multi-class classifications often fall short, further impairing model performance \cite{Abedi20233535}. This complexity is underscored by Liao et al. \cite{Liao20216609}, who show that even advanced models like Deep Facial Spatiotemporal Network (DFSTN) achieve less than $60\%$ accuracy on multi-class engagement tasks in \textit{DAiSEE}. As noisy labels degrade generalization and trigger overfitting, Zhong et al. \cite{Zhong20221290} underscore the need for effective noise-handling strategies. Jiang and Deng \cite{Jiang20232402} highlight that without these approaches, emotion recognition performance suffers.

\subsection{Learning with Noisy Labels}
In the literature, samples whose observed labels match the true labels are called ‘clean’ and those with corrupted or incorrect labels ‘noisy’ \cite{arazo2019unsupervised}. Surveys by Song et al. \cite{song2022learn} and Han et al. \cite{han2021survey} provide comprehensive taxonomies of these definitions. Noisy labels can cause Deep Neural Networks (DNNs) to overfit, leading to suboptimal performance. While many LNL techniques, including module-based, loss-based, label correction, and sample selection methods, have been proposed to enhance generalization, their application in engagement analysis remains unexplored \cite{northcutt2021pervasivelabelerrorstest}. Therefore, we examine methods from a broader emotion recognition scope.

\vspace{.05in} \noindent{\textbf{Module-based methods}} modify the neural network modules to be more robust against label noise. In depression detection, Chen et al. \cite{Chen20222106} leverage soft labels and attention to design a module-based method that enhances neural network generalization despite label noise. Tan et al. \cite{Tan2024} present RCL-Net, a module-based approach that uses dual-consistency learning to improve representation quality under noisy label conditions in facial expression datasets. Building on earlier strategies, Liu et al. \cite{Liu_10214365} integrated auxiliary tasks and graph-based semantic reasoning, which significantly improved robustness to noisy labels in facial expression recognition. Subsequently, Mao et al. \cite{Mao2022818} introduced DMM-CNN, which integrates task-specific features with dynamic weighting and adaptive thresholding to boost generalization in subjective, imbalanced facial attribute classification tasks.

\vspace{.05in} \noindent{\textbf{Loss-based methods}} design the loss function to be more robust against label noise. Khan et al. \cite{khan2023novellossfunctionutilizing} tackle subject-dependent noise by introducing a Wasserstein Distance-based loss function, thus improving model generalization in a range of emotion recognition applications. Further enhancing loss-based approaches, Jiang and Deng \cite{Jing_10321660} demonstrated that attention mechanisms such as Attention Flipping Consistency (AFC) and label-guided spatial attention dispersing (SAD) losses improve robustness by guiding the model to concentrate on distinct, consistent expression features, mitigating label ambiguity. Chen et al. \cite{Chen_2024_static} further advanced loss-based methods by proposing an Emotion-Anchors Self-Distillation Loss, which reduces label ambiguity and enhances dynamic facial expression recognition.

\vspace{.05in} \noindent{\textbf{Label correction methods}} adjust the loss value or correct the labels to mitigate the impact of label noise. Zhou et al. \cite{Zhou_PR_PL_10160130} address label noise in EEG emotion datasets with their Prototypical Representation-based Pairwise Learning (PR-PL) framework, which employs adaptive pseudo-labeling and pairwise similarity learning to boost robustness via domain alignment. Washington et al. \cite{Washington20211363} revealed that crowdsourced soft-target labels capture the nuanced subjectivity of emotions, which improves model robustness and mirrors human judgment. Jin et al. \cite{Jin20247395} adopted Gaussian Mixture Models to generate soft pseudo-labels, effectively reducing label noise and elevating mental state classification accuracy in social media text. Liu et al. \cite{Liu_2022} introduced ULC-AG, a method that integrates graph-based facial action unit modeling with semantic similarity to effectively correct noisy labels in extensive facial expression datasets. Zhang and Etemad \cite{zhang2023partiallabellearningemotion} showed that Partial Label Learning with label disambiguation can effectively refine ambiguous EEG emotion labels, mitigating label noise. Complementary to these methods, several recent approaches exploit multimodal large language models (MLLMs) and VLMs as external teachers. Huang et al. \cite{huang2024mmultimodal} formulate “Machine Vision Therapy”, where an MLLM rectifies vision model predictions and improves Out-of-Distribution (OOD) robustness via denoising in-context learning; Wang et al. \cite{NEURIPS2024wang} propose NoiseGPT, which detects and cleanses label noise by analyzing probability curvature and in-context discrepancies; and Zhang et al. \cite{zhang2024adaneg} introduce AdaNeg, a training-free OOD detector that constructs adaptive negative proxies from test-time feature banks. Our work similarly leverages a VLM as a teacher, but differs in using offline, questionnaire-guided behavioral cues to assign reliability scores and refine subjective engagement labels, which then drive curriculum learning and soft-label training instead of modifying predictions on-the-fly. 

Our key methodological contribution is to decouple label reliability from semantic content using a questionnaire and a VLM, and to exploit this reliability in a two-stage curriculum with ambiguity-aware soft labels for adjacent ordinal classes. Unlike prior VLM-based denoising or OOD work, we apply reliability-aware training to subjective engagement and explicitly organize learning by estimated reliability.

\vspace{.05in} \noindent{\textbf{Sample selection methods}} choose clean samples from noisy datasets to improve the model’s generalization performance. Gera and Balasubramanian \cite{Darshan_9607483} developed Consensual Collaborative Training, a method that emphasizes clean samples during early training while enforcing consistency loss to improve learning across noisy datasets. Building on previous methods, Chen et al. \cite{chen_13193891} introduced SSPA-JT, a technique that leverages clean sample filtering and paired augmentation to dynamically balance the dataset and improve generalization in facial expression recognition. Jiang and Deng \cite{Jiang20232402} advanced sample selection by introducing the Progressive Teacher framework, which iteratively discards high-loss samples, improving generalization in noisy facial expression recognition.

\subsection{VLMs in Affective Computing}
Affective Computing broadly investigates computational approaches for sensing, interpreting, and modelling human emotions and related states in context \cite{kumar2024nontypical}. In this work, we treat students' engagement as a non-typical affective state and use VLMs as tools to improve its annotation and recognition under subjective labeling.

VLMs have set new benchmarks in multi-modal tasks, particularly in image captioning and visual question answering \cite{shvetsova2024how}. By leveraging lightweight adapters linking visual and language models, they extend their potential to Affective Computing, where the fusion of text, visuals, audio, EEG data helps address challenges like micro-expression recognition, sentiment analysis, brain-computer interactions \cite{zhang2024AC}.

Until now, very few works have utilized VLMs in Affective Computing. Recent empirical evaluations of Teotia et al. and Wang et al. consider VLMs for learning engagement detection and academic emotion analysis, treating the models as direct predictors of student states \cite{Teotia10917661, wang2025usingvisionlanguagemodels}. The results underline significant room for improvement, particularly in handling nuanced and context-dependent engagement cues. Xie et al. \cite{xie2024emov} presented EmoVIT using GPT-assisted visual tuning for enhanced emotion reasoning, but it does not address subjectivity. Nadeem et al. \cite{Nadeem20242566} found VLMs to perform similarly with deep models on typical emotion datasets, restricting their utility for non-typical emotions. While AU-LLaVA \cite{hu2024unified} and GPT-4V \cite{lu2024gptp} excel in tasks like Action Unit recognition, their limited focus hinders broader applicability. Etesam et al. \cite{etesam2024cont} achieved notable results with image captioning, yet static prompts reduce flexibility in handling subjective non-typical emotions.

Prior applications of VLMs and LNL techniques in Affective Computing cover both typical emotions and initial engagement studies, but mostly rely on simple prediction pipelines with little support for label noise or subjectivity. This highlights the need for VLM-driven strategies that explicitly improve dataset reliability and learning in subjective, non-typical states.

\section{Subjective Labels and VLM Filtering}\label{sec:3_method}
\subsection{Datasets}\label{sec:VLM_datasets} Three engagement datasets are utilized in the course of this study: \textit{EngageNet} \cite{singh2023do, dhall2023emotiw}, \textit{PAFE} (Predicting Attention with Facial Expression) \cite{Taeckyung2022PAFE}, and \textit{DREAMS} (Diverse Reactions of Engagement and Attention Mind States) \cite{Singh2024dreams}. Following the original papers, we report classification accuracy and weighted F1, matching their evaluation protocols. Performance is always measured within each dataset’s original engagement scale and video window. Datasets were licensed, anonymized per GDPR, and bias-evaluated via VLMs.

\vspace{0.05in} \noindent{\textbf{\textit{EngageNet}}} 
dataset comprises $11,300$ video samples from $127$ participants aged $18$ to $37$, interacting with a web-based learning platform. Video samples have been annotated for four levels of engagement: `Highly Engaged,' `Engaged,' `Barely Engaged,' `Not Engaged.' Dataset includes three subject-independent sets: $7,983$ training videos from $90$ participants, $1,071$ validation videos from $11$ participants, $2,257$ test videos from $26$ participants. Each video sample was independently annotated by three experts.

\vspace{0.05in}\noindent{\textbf{\textit{PAFE}}} dataset captures $1,100$ video samples from $15$ university students attending an online lecture. Each sample was recorded at $640$p resolution and $30$fps, with probes conducted every $40$ seconds to label attentional states as `Focused,' `Not-Focused,' or `Skip.' Probes were annotated by participants’ self-reports and expert review to remove drowsy mismatches.

\vspace{0.05in} \noindent{\textbf{\textit{DREAMS}}} dataset is built from video recordings of $32$ participants captured in natural environments. In total, $832$ video segments of $40$ seconds each were recorded, with $781$ segments successfully used for analysis. Participants provided self-ratings of their engagement while watching a range of video clips that were chosen to evoke different emotional reactions. The data is divided into subject-independent sets. Every $40$ seconds, participants annotated their engagement levels via self-assessment prompts.

\vspace{.08in}
\noindent The details of the computational settings used in this work are in Appendix~\ref{sec:Computational_Settings}.

\subsection{VLM-based Engagement Classification}\label{sec:VLM_analysis} We leverage several state-of-the-art open-source VLMs (including InternVL$2$ ($8$B and $26$B) \cite{chen2023internvl, chen2024far}, LLaVA $1.6$ ($7$B) \cite{liu2024llavanext, liu2023improvedllava, liu2023llava}, and VILA $1.5$ ($8$B) \cite{lin2023vila}) to address a four-class engagement classification task on the \textit{EngageNet} dataset using zero- and few-shot learning approaches. Due to privacy constraints, proprietary VLMs were not used, as the sensitive video samples cannot be uploaded externally.

\vspace{0.05in} \noindent{\textbf{Single question mode.}} We evaluated three input types: single images, multiple images, and video sequences (extracted frames). For multiple images, the model predicts the engagement level for each frame independently, and the final prediction is the mode of these individual predictions.

\vspace{0.05in} \noindent{\textbf{Conversation mode.}} To enhance VLM discriminative power, we evaluated InternVL$2$ in conversation mode with single image and video inputs, using both $8$B and $26$B model sizes. InternVL$2$ was chosen for its Conversation mode (absent in LLaVA) and its strong performance in single-question settings. In the `few-shot/image' mode, the model was first provided with a few example images for each engagement class along with corresponding affective descriptions. Subsequently, it was prompted to analyze a new image and finally output a numeric engagement level. In the zero-shot video approach, the model is provided with a video from which it extracts key frames (eight by default) and, through sequential conversational rounds, is instructed first to focus on specific video features and then to classify the engagement level using the same methodology as in the few-shot/image mode. Detailed prompt formulations are provided in Appendix~\ref{sec:Conversation_Mode_Prompt}.

\vspace{0.05in} \noindent{\textbf{VLM guidance with structured prompts.}} Without clear, direct instructions defining engagement states, VLM predictions in both Single Question and Conversation modes can suffer from inconsistency and reduced accuracy. To address this, we adopted a structured approach using precise YES/NO prompts that bind the model's responses to specific, observable behaviors. In our framework, the VLM is guided by a questionnaire (see Appendix~\ref{sec:Questionnaire_Details}) organized into four key categories: \textbf{\textit{(1) Eye Gaze and Attention}}, \textbf{\textit{(2) Facial Expressions}}, \textbf{\textit{(3) Body Language and Posture}}, \textbf{\textit{(4) Distractions and External Activities}}. This approach allows to investigate whether behavior-based annotation reduces inter-annotator variability and helps the VLM avoid learning annotator-specific biases instead of generalizable engagement features.

The resulting checklist is deliberately static and behavior-grounded: items capture gaze, posture, facial and distraction cues commonly used in human engagement protocol \cite{singh2023do, dhall2023emotiw,Taeckyung2022PAFE,Singh2024dreams}, selected to balance coverage with simplicity. We apply this fixed template across datasets instead of searching for an “optimal” questionnaire, as our goal is to analyse VLM-guided label quality rather than optimize the question set. The full template is given in Appendix~\ref{sec:Questionnaire_Details} for reuse and future domain-specific refinement.

\subsection{Enhancing Data Quality with VLM Filtering}\label{sec:VLM_filtering}
We leverage VLM outputs not only for engagement classification but also to assess label reliability by partitioning the Validation sets of \textit{EngageNet} and \textit{PAFE} datasets into two subsets: Accepted (where VLM predictions agree with the ground truth) and Rejected (where discrepancies exist). Following the clean/noisy nomenclature, we treat our Accepted subset as a proxy for clean data and Rejected as a proxy for noisy data. Unlike standard noise-model approaches (which consider each noisy label as inversion from an assumed ground truth) we use VLM agreement as a model-based measure of sample reliability. To assess this, we evaluate classical computer vision models (TCCT-Net \cite{Vedernikov_2024_CVPR}, ResNet+TCN \cite{abedi2021improving}, EfficientNet+LSTM \cite{selim2022students}, EfficientNet+Bi-LSTM \cite{selim2022students}) across the full Validation set as well as the Accepted and Rejected subsets. This comparative analysis reveals the models’ sensitivity to label noise and highlights their reliance on high-consensus annotations. The filtering criteria and experimental protocol are discussed in Appendix~\ref{sec:Enhancing_Data} and ~\ref{sec:cv_classic}.

From a computational perspective, VLM-based filtering is run once as an offline preprocessing stage (see Appendix~\ref{sec:Computational_Settings} for hardware details), and the resulting Accepted/Rejected indicators and reliability scores are cached for all subsequent analyses and curriculum runs. This step scales roughly linearly with the number of clips, while the per-epoch training throughput of our engagement models stays unchanged. For larger datasets or heavier VLMs, the cost can be further reduced via batched video processing and mixed-precision (e.g., FP$16$) inference, but a detailed efficiency study is beyond the scope of this work.

\subsection{Human Performance in Engagement Analysis}
To further assess ground truth label reliability, we conducted a human evaluation comparing annotator consistency with VLM performance, highlighting alignment and ambiguity in subjective engagement classification. Ten experienced annotators classified $50$ video samples from \textit{EngageNet} Validation set into four engagement levels. Annotators received detailed guidelines and were allowed up to two views per video. To account for minor discrepancies, we introduced a tolerance-based accuracy metric (±$1$ level). Full experimental details are in Appendix~\ref{sec:Human_Performance_Appendix}. 

\section{Baseline Results: VLM Classification and Data Filtering}\label{sec:4_experiments}
\subsection{VLM-based Engagement Classification}
Table~\ref{tab:starting_comparison_acc} of Appendix~\ref{sec:vlm_eng_class} presents the zero- and few-shot performance of VLMs on the \textit{EngageNet} validation set. While LLaVA achieved $40.06\%$ accuracy in Single Question mode using two images, it lacks conversation mode support. In contrast, InternVL$2$ improved from $28.66\%$ (Single Question mode) to $35.20\%$ in Conversation mode, and further reached $45.28\%$ with structured question guidance and video input using the $8$B model (vs. $29.97\%$ for the $26$B variant). Temporal and sequential gains motivated our use of InternVL$2$ ($8$B) for subsequent experiments.

\subsection{Evaluation of VLM Filtering and its Impact on Model Performance}
Structured question guidance (Table~\ref{tab:engagement_questions}, Appendix~\ref{sec:Human_Performance_Appendix}) enabled VLM zero-shot video classification with InternVL$2$ ($8$B) to reach $45.28\%$ accuracy on full Validation set of \textit{EngageNet} (thus, resulting in $485$ accepted, $586$ rejected) and $71.47\%$ on full Validation set of \textit{PAFE} (providing $243$ accepted, $97$ rejected). This highlights VLM’s capacity to detect annotation inconsistencies, providing refined data for assessing classical models’ overfitting to noisy labels.

\begin{table}[h]
    \centering
    \setlength\tabcolsep{2pt}
    \resizebox{0.85\textwidth}{!}{
    \begin{tabular}{@{}lccc@{}}
        \toprule
        \textbf{Method} & \textbf{Full Validation} & \textbf{Rejected} & \textbf{Accepted} \\
        \midrule
        \multicolumn{4}{c}{\textit{EngageNet Dataset}} \\
        \midrule
        VLM -- InternVL2 (8B) \cite{chen2023internvl,chen2024far} & 45.28 & -- & -- \\
        TCCT-Net \cite{Vedernikov_2024_CVPR} & \underline{68.91} & 58.02 \textcolor{red}{\fontsize{5}{6}\selectfont$-10.89$} & \textbf{82.06} \textcolor{apricot}{\fontsize{5}{6}\selectfont$+13.15$} \\
        ResNet + TCN \cite{abedi2021improving} & \underline{54.72} & 46.65 \textcolor{red}{\fontsize{5}{6}\selectfont$-8.07$} & \textbf{64.47} \textcolor{apricot}{\fontsize{5}{6}\selectfont$+9.75$} \\
        EfficientNet + LSTM \cite{selim2022students} & \underline{57.57} & 49.37 \textcolor{red}{\fontsize{5}{6}\selectfont$-8.20$} & \textbf{67.48} \textcolor{apricot}{\fontsize{5}{6}\selectfont$+9.91$} \\
        EfficientNet + Bi-LSTM \cite{selim2022students} & \underline{58.94} & 51.17 \textcolor{red}{\fontsize{5}{6}\selectfont$-7.77$} & \textbf{68.33} \textcolor{apricot}{\fontsize{5}{6}\selectfont$+9.39$} \\
        \midrule
        \multicolumn{4}{c}{\textit{PAFE Dataset}} \\
        \midrule
        VLM -- InternVL2 (8B) \cite{chen2023internvl,chen2024far} & 71.47 & -- & -- \\
        TCCT-Net \cite{Vedernikov_2024_CVPR} & \underline{74.71} & 17.53 \textcolor{red}{\fontsize{5}{6}\selectfont$-57.18$} & \textbf{95.88} \textcolor{apricot}{\fontsize{5}{6}\selectfont$+21.17$} \\
        ResNet + TCN \cite{abedi2021improving} & \underline{62.37} & 31.52 \textcolor{red}{\fontsize{5}{6}\selectfont$-30.85$} & \textbf{74.68} \textcolor{apricot}{\fontsize{5}{6}\selectfont$+12.31$} \\
        EfficientNet + LSTM \cite{selim2022students} & \underline{65.23} & 35.68 \textcolor{red}{\fontsize{5}{6}\selectfont$-29.55$} & \textbf{77.03} \textcolor{apricot}{\fontsize{5}{6}\selectfont$+11.80$} \\
        EfficientNet + Bi-LSTM \cite{selim2022students} & \underline{67.12} & 36.13 \textcolor{red}{\fontsize{5}{6}\selectfont$-30.99$} & \textbf{79.49} \textcolor{apricot}{\fontsize{5}{6}\selectfont$+12.37$} \\
        \bottomrule
    \end{tabular}}
    \caption{Evaluation of classical computer vision model accuracy across Validation, Accepted, and Rejected subsets in \textit{EngageNet} and \textit{PAFE} (all values represent percentages). \textbf{Bold} indicates the best performance and \underline{underline} shows the second-best.}
    \label{tab:traditional_cv}
\end{table}

\noindent{\textbf{Classical computer vision results.}} Table~\ref{tab:traditional_cv} reveals a strong dependence on label quality.  On the \textit{EngageNet} dataset, TCCT-Net achieves $82.06\%$ accuracy on Accepted subset, but this drops by $24.04\%$ to $58.02\%$ on Rejected subset. Similarly, on the \textit{PAFE} dataset, TCCT-Net’s accuracy falls from $95.88\%$ on Accepted samples to $17.53\%$ on Rejected ones with a decline of $78.35\%$. Such disparities indicate that aggregate Validation scores (e.g., $68.91\%$ for TCCT-Net on \textit{EngageNet} and $74.71\%$ on \textit{PAFE}) may mask the underlying sensitivity of models to label noise. Overall, these results (also observed for ResNet+TCN and EfficientNet-based models) underscore that even minor label inconsistencies impair performance, emphasizing the need for robust data curation and noise mitigation in engagement analysis. Table~\ref{tab:traditional_cv} analyses only expert-labeled \textit{EngageNet} and \textit{PAFE} (expert-confirmed after self-reports), while \textit{DREAMS} dataset (built entirely on participant self‐ratings) is discussed in Section~\ref{sec:exp_results}.

\subsection{Human Performance Results}
Human evaluation, ground truth annotation and VLM results on the \textit{EngageNet} Validation set revealed subjectivity in engagement classification. Human evaluators achieved $46.25\%$ exact-match accuracy, with $42.00\%$ on Accepted samples and $50.50\%$ on Rejected ones, suggesting VLM-filtered ``reliable" samples remain ambiguous due to labeling variations. However, when applying ±$1$ tolerance metric, human accuracy improved dramatically to $88.75\%$ (with $82.50\%$ on Accepted and $95.00\%$ on Rejected samples), indicating that while exact matches are challenging, annotators generally select engagement levels close to ground truth. In comparison, the VLM achieved an overall accuracy of $50.70\%$ and ±$1$ tolerance accuracy of $78.52\%$, demonstrating that when guided by structured prompts, it can match or even surpass human consistency. Findings show that ground truth labels fail to capture a nuanced nature of engagement: low exact-match accuracies and higher tolerance-based scores. This underscores the need for improved annotation protocols and model designs that better accommodate subjectivity in engagement classification.

\begin{table}[h]
    \centering
    {\fontsize{8.1}{9.5}\selectfont
    \setlength\tabcolsep{4pt}
    \resizebox{0.85\textwidth}{!}{\begin{tabular}{@{}c c c c c c@{}}
        \toprule
        \makecell{\textbf{Frames /}\\\textbf{Questions}} & \makecell{\textbf{Full Validation}\\\textbf{Accuracy (\%)}} & \multicolumn{2}{c}{\makecell{\textbf{Rejected}\\\textbf{Accuracy (\%) /}\\\textbf{Samples}}} & \multicolumn{2}{c}{\makecell{\textbf{Accepted}\\\textbf{Accuracy (\%) /}\\\textbf{Samples}}} \\
        \midrule
        \textbf{F2}  & 43.04 & 62.13 \textcolor{red}{\fontsize{5}{6}\selectfont$-6.78$} & 610 & \textbf{77.87} \textcolor{apricot}{\fontsize{5}{6}\selectfont$+8.96$} & 461 \\
        \textbf{F16} & 50.70 & 50.19 \textcolor{red}{\fontsize{5}{6}\selectfont$-18.72$} & 528 & \textbf{87.11} \textcolor{apricot}{\fontsize{5}{6}\selectfont$+18.20$} & 543 \\
        \textbf{F20} & 47.99 & 52.78 \textcolor{red}{\fontsize{5}{6}\selectfont$-16.13$} & 557 & \textbf{86.38} \textcolor{apricot}{\fontsize{5}{6}\selectfont$+17.47$} & 514 \\
        \midrule
        \textbf{Q3}  & 36.23 & \textbf{69.84} \textcolor{apricot}{\fontsize{5}{6}\selectfont$+0.93$} & 683 & 67.27 \textcolor{red}{\fontsize{5}{6}\selectfont$-1.64$} & 388 \\
        \textbf{Q9}  & 43.79 & 55.48 \textcolor{red}{\fontsize{5}{6}\selectfont$-13.43$} & 602 & \textbf{86.14} \textcolor{apricot}{\fontsize{5}{6}\selectfont$+17.23$} & 469 \\
        \textbf{Q15} & 50.70 & 50.19 \textcolor{red}{\fontsize{5}{6}\selectfont$-18.72$} & 528 & \textbf{87.11} \textcolor{apricot}{\fontsize{5}{6}\selectfont$+18.20$} & 543 \\
        \bottomrule
    \end{tabular}}}
    \caption{Performance of VLM over different frame counts (F) and prompt question (Q) counts. Red/green values indicate accuracy differences from the TCCT-Net \cite{Vedernikov_2024_CVPR} baseline of $68.91\%$. Total number of samples in the Validation set of \textit{EngageNet} is $1071$. \textbf{Bold} indicates the best performance. Detailed results are provided in Table~\ref{tab:ablation_merged} of Appendix~\ref{sec:Ablation_Test}.}
    \label{tab:ablation_merged_short}
\end{table}

\subsection{Ablation Study}
To assess the impact of input configuration on engagement classification, we did ablation studies on the number of input frames and prompt questions in VLM filtering (Table \ref{tab:ablation_merged_short}). Complete results for all configurations are provided in Table~\ref{tab:ablation_merged} of Appendix~\ref{sec:Ablation_Test}. Experiments reveal that increasing the number of frames improves performance, with accuracy peaking at $16$ frames. Similarly, varying the number of prompt questions, we observed that using all $15$ questions yields the best performance, as it provides richer set of engagement cues for VLM filtering. Results indicate optimal balance between temporal context and prompt detail, critical for mitigating effects of subjective labeling.

\section{Curriculum Learning with Soft Label Refinement}\label{sec:curriculum}
Due to label inconsistencies in Sections~\ref{sec:3_method} and \ref{sec:4_experiments}, we propose a training strategy integrating curriculum learning with soft label refinement, guided by labels derived from VLM analysis. Gradually incorporating challenging samples and using soft labels, our approach mitigates label noise and subjectivity. Validation set evaluations confirm its effectiveness. This framework is implemented to improve TCCT-Net, a two-stream architecture for fast engagement analysis by Vedernikov et al. \cite{Vedernikov_2024_CVPR}, and is adaptable to other models. We select TCCT-Net as our main backbone because it is a recent, efficient engagement estimator that already demonstrates strong performance on the benchmarks considered in this work.

\subsection{Curriculum Learning Strategy}
\begin{figure}[h]
    \centering
    \includegraphics[width=0.6\linewidth]{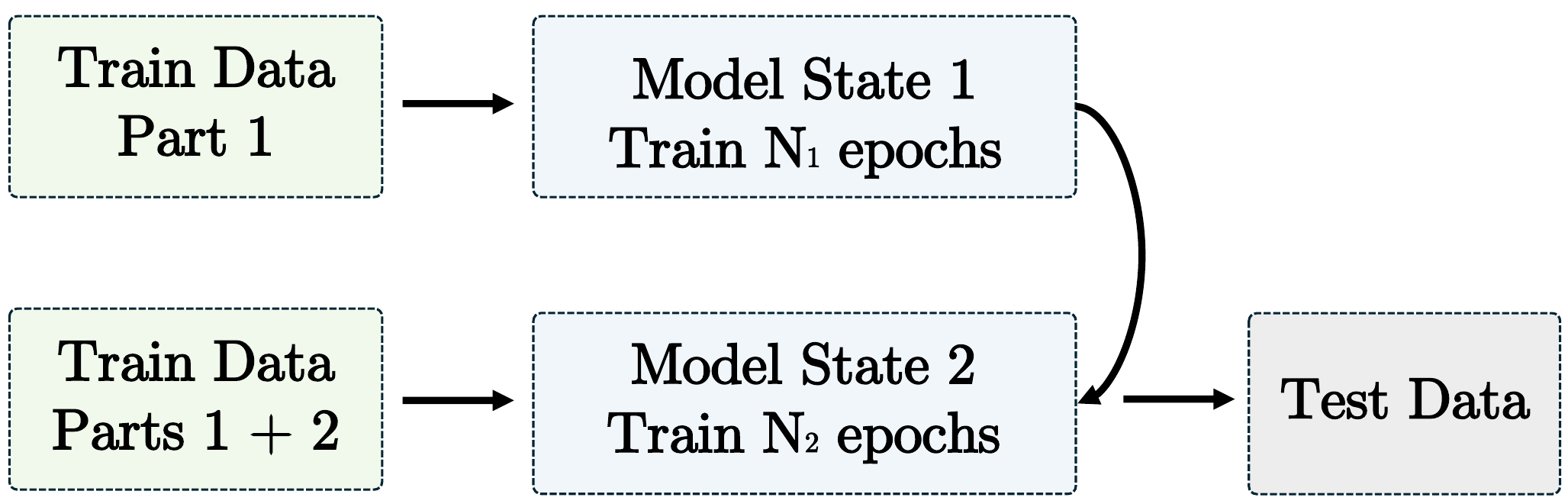}
    \caption{Two-stage curriculum learning scheme for engagement estimation: Stage $1$ uses high-confidence samples; Stage $2$ incorporates ambiguous samples.}
    \label{fig:CL_scheme}
\end{figure}

Our curriculum learning strategy (Figure~\ref{fig:CL_scheme}) is built on the idea that not all training samples are equal, some are more reliable than others. To that end, we rank and split the training data based on the difference between the ground truth (GT) label and the VLM-predicted label, which serves as a proxy for sample reliability. We treat samples with zero or with differences of $2$ or $3$ (typically VLM errors) as high-confidence, using them in the first training stage, and reserve samples with a one-level difference for a later stage due to their subjectivity. This design reflects the ordinal nature of engagement: adjacent classes ($|y-\hat{y}| = 1$) differ only slightly in intensity and are therefore more subjective. In contrast, larger gaps ($|y-\hat{y}| \geq 2$) indicate clearly distinct engagement regimes; we interpret them mainly as VLM failures and trust the human label, treating these samples as high-confidence in the first stage. By training first on reliable samples and only later on ambiguous ones, the model learns more robust features and generalizes better. We explicitly do not treat VLM agreement as absolute truth; it is merely a proxy for ambiguity that guides the optimizer from the least ambiguous to the most difficult samples.

\subsection{Soft Label Generation and Integration}
To further mitigate the impact of noisy or subjective labels, we generate soft labels for samples with minor discrepancies (one-level differences), converting hard one-hot vectors into probability distributions that express uncertainty. This integration into the loss function mitigates the negative effects of mislabeling.

Let \( y \in \{0,1,2,3\} \) denote the ground truth label and \(\hat{y}\) the VLM-predicted label. We define the soft label vector \( \mathbf{s} \in \mathbb{R}^4 \) as:

\begin{equation}
    s_i =
    \begin{cases}
        \alpha, & \text{if } |y-\hat{y}| = 1 \text{ and } i = y, \\
        1 - \alpha, & \text{if } |y-\hat{y}| = 1 \text{ and } i = \hat{y}, \\
        1, & \text{if } |y-\hat{y}| \neq 1 \text{ and } i = y, \\
        0, & \text{otherwise},
    \end{cases}
    \label{eq:1}
\end{equation}
where \( \alpha \in (0,1) \) controls confidence in the GT versus the VLM prediction. In practice, we set the smoothing coefficient $\alpha$ through a grid search. This formulation applies soft labels (i.e., probability distributions) to samples with a one-level discrepancy, while reverting to standard one-hot encoding for unambiguous cases.
The integration of soft labels into training is achieved by modifying the loss function. For ambiguous samples, i.e. a one-level discrepancy (\(|y-\hat{y}| = 1\)), we use a soft loss defined via Kullback–Leibler (KL) divergence:

\begin{equation}
    \mathcal{L}_{\text{soft}} = \sum_{i=0}^{3} s_i \log \frac{s_i}{p_i},
    \label{eq:2}
\end{equation}
where \(p_i = \text{softmax}(z_i)\) is the predicted probability for class \(i\). For unambiguous samples, we apply the standard cross-entropy loss:

\begin{equation}
    \mathcal{L}_{\text{hard}} = -\log(p_y).
    \label{eq:3}
\end{equation}
The per-sample loss is therefore:

\begin{equation}
    \mathcal{L} =
    \begin{cases}
        \mathcal{L}_{\mathrm{soft}}(\mathbf{s},p), & \text{if } |y-\hat{y}| = 1, \\
        \mathcal{L}_{\mathrm{hard}}(p_{y}), & \text{otherwise},
    \end{cases}
    \label{eq:4}
\end{equation}
with total training loss computed as:
\begin{equation}
    \mathcal{L}_{\text{total}} = \frac{1}{N}\sum_{j=1}^{N} w_j\,\mathcal{L}_j,
    \label{eq:5}
\end{equation}
where \(w_j\) reflects sample reliability and \(N\) is the batch size. 

This loss function enables the network to learn from ambiguous samples, reducing overfitting to hard labels by incorporating insights from both GT and VLM. Although our soft targets come from VLM-GT discrepancies rather than multi-annotator distributions, they nevertheless inject a useful uncertainty signal that stabilizes training.

\subsection{Weighted Data Augmentation via Segmentation and Recombination}
To incorporate sample reliability into training, we modify the standard Segmentation and Recombination (S\&R) augmentation, originally used in TCCT-Net, by including sample weights. Each sample is assigned a weight \(w\) reflecting its label reliability, defined by the discrepancy between the GT and VLM prediction. Weights are used to compute a sampling probability, ensuring that segments from more reliable samples are more likely to be selected, and they determine the average weight of each augmented example.

Let the dataset be \(\{(x_i, y_i, w_i)\}_{i=1}^{N}\), where \(x_i \in \mathbb{R}^{L}\) is a signal of length \(L\), \(y_i\) its label, and \(w_i\) its weight. Each signal is segmented into \(n\) equal parts of length \(l = L/n\). For class \(c\), denote its sample indices by \(\mathcal{I}_c = \{ i \mid y_i = c \}\). Sampling probability for each sample is defined as:
\begin{equation}
    p_i = \frac{w_i}{\sum_{j \in \mathcal{I}_c} w_j}.
    \label{eq:6}
\end{equation}
To form an augmented sample \(x_{\text{aug}}\in\mathbb{R}^{L}\) for class \(c\), we:

\begin{enumerate}
    \item For each segment \(j\) (\(1\leq j\leq n\)), randomly select a sample index \(i_j \in \mathcal{I}_c\) with probability \(p_{i_j}\).\vspace{.03in}
    \item Extract the \(j\)th segment as \(x_{i_j}^{(j)} = x_{i_j}[(j-1)l : jl]\).\vspace{.05in}
    \item Concatenate segments \(x_{\text{aug}} = [x_{i_1}^{(1)}, x_{i_2}^{(2)}, \dots, x_{i_n}^{(n)}]\).\vspace{.07in}
    \item Compute the augmented weight as \(w_{\text{aug}} = \frac{1}{n}\sum_{j=1}^{n} w_{i_j}\).\vspace{.05in}
\end{enumerate}

This strategy ensures that segments from more reliable samples (with higher \(w_i\)) are preferentially selected, preserving signal integrity and reducing noise. The augmented data \(\{(x_{\text{aug}}, c, w_{\text{aug}})\}\) is then merged with the original dataset, increasing effective dataset size while integrating label confidence into training. Modified S\&R augmentation provides data diversification along with a refined weighting mechanism reflecting label reliability.

\subsection{VLM-guided Label Refinement}
VLM output serves as a critical guide for quantifying label noise and informing training stages. The discrepancy between the GT and VLM-predicted labels provides an objective measure of reliability informing sample ranking in curriculum learning strategy and the generation of soft labels. By integrating VLM guidance into sample ordering and label adjustment, our method selectively weights training examples and mitigates the effects of noisy annotations.

Curriculum learning, by starting with high-confidence, low-noise samples \cite{bengio2009curriculum}, establishes a solid foundation for handling ambiguous data. Soft label refinement smooths the supervision signal, decreasing sensitivity to errors and enhancing generalization \cite{szegedy2016rethinking}. Leveraging VLM insights motivated by ensemble and co-training paradigms \cite{zhou2012ensemble} further mitigates overfitting and improves accuracy.

Together, these components enable the network to adapt effectively to ambiguous samples, ultimately leading to improved generalization and robustness on unseen test data. However, the agreement between VLM and human labels is only used as a proxy for sample reliability and does not imply that the VLM predictions are always correct. In our design, (i) VLM–human agreement and large discrepancies are used only to stage the curriculum and define sample weights, (ii) one-level disagreements are encoded as $\alpha$-weighted soft targets that still preserve the original ground-truth label, and (iii) samples never have their human labels replaced by VLM predictions. As shown in Section~\ref{sec:4_experiments}, VLM accuracy is prompt- and setting-dependent, and can vary substantially, so relying on it blindly would be inappropriate. A formal convergence analysis under VLM-induced noise is outside the present scope, and we position the method as an empirically validated effective scheme whose formal analysis is left to future work.

\begin{table}[!h]
    \centering
    \scriptsize
    \setlength{\tabcolsep}{5pt}
    \label{tab:combined_results}
    \begin{subtable}{\columnwidth}
        \centering
        \setlength{\tabcolsep}{2.4pt}
        \resizebox{\textwidth}{!}{%
            \begin{tabular}{l c c c c c c}
                \toprule
                \makecell{\textbf{Feature}} &
                \makecell{\textbf{Transformer}} &
                \makecell{\textbf{LSTM}} &
                \makecell{\textbf{TCN}} &
                \makecell{\textbf{CNN-LSTM}} &
                \makecell{\textbf{TCCT-Net}} &
                \makecell{\textbf{TCCT-Net}\\\textbf{+ VLM}} \\
                \midrule
                \textbf{EG}      & 55.45 & 61.34 & 63.03 & 60.78 & \underline{64.33} & \textbf{64.61} \\
                \textbf{HP}      & --  & 67.41 & 67.51 & 67.88 & \textbf{68.91} & \underline{68.16} \\
                \textbf{AU}      & -- & 62.75 & 64.24 & 62.00 & \textbf{66.29} & \underline{66.01} \\
                \textbf{EG + HP} & 64.45  & 65.17 & -- & -- & \textbf{65.64} & \underline{65.55} \\
                \textbf{HP + AU}      & -- & -- & -- & -- & \underline{65.08} & \textbf{66.29} \\
                \textbf{AU + EG}      & -- & -- & -- & -- & \underline{65.17} & \textbf{65.73} \\
                \bottomrule
            \end{tabular}%
        }
        \caption{\textbf{EngageNet}: Classification accuracies (\%) for Transformer, LSTM, TCN, and CNN-LSTM from \cite{singh2023do}; TCCT-Net from \cite{Vedernikov_2024_CVPR}.}
        \label{tab:results_engagenet}
    \end{subtable}
    \vspace{1em}

    \begin{subtable}{\columnwidth}
        \centering
        \setlength{\tabcolsep}{2.4pt}
        \resizebox{\textwidth}{!}{%
            \begin{tabular}{l c c c c c}
                \toprule
                \makecell{\textbf{Feature}} &
                \makecell{\textbf{ST}} &
                \makecell{\textbf{MT}} &
                \makecell{\textbf{TL}} &
                \makecell{\textbf{TCCT-Net}} &
                \makecell{\textbf{TCCT-Net}\\\textbf{+ VLM}} \\
                \midrule
                \textbf{EG}  & 0.185 & 0.185 & 0.185 & \underline{0.357} & \textbf{0.423}  \\
                \textbf{HP}      & --  & -- & -- & \underline{0.185} & \textbf{0.405}  \\
                \textbf{AU}      & -- & -- & -- & \underline{0.316} & \textbf{0.343}  \\
                \textbf{EG + HP}     & 0.181 & \underline{0.218} & 0.165 & 0.185 & \textbf{0.434}  \\
                \textbf{HP + AU}      & -- & -- & -- & \underline{0.185} & \textbf{0.316} \\
                \textbf{EG + AU}      & -- & -- & -- & \underline{0.339} & \textbf{0.386} \\
                \textbf{EG + HP + AU}      & 0.256 & \underline{0.306} & 0.216 & 0.185 & \textbf{0.384} \\
                \textbf{EG + HP + AU + LMK}     & \underline{0.226} & 0.208 & \textbf{0.293} & -- & -- \\
                \textbf{EG + HP + AU + LMK + PDM}    & \textbf{0.305} & \underline{0.299} & 0.284 & -- & -- \\
                \bottomrule
            \end{tabular}%
        }
        \caption{\textbf{DREAMS}: Weighted F1 scores for various feature combinations. Results for Single-task (ST), Multi-task (MT), and Transfer Learning (TL) from \cite{Singh2024dreams}; TCCT-Net from \cite{Vedernikov_2024_CVPR}.}
        \label{tab:results_dreams}
    \end{subtable}
    \vspace{0.6em}

    \captionsetup[subtable]{skip=4pt}
    \begin{subtable}{\columnwidth}
        \centering
        \scriptsize
        \scalebox{1.1}{%
            \begin{tabular}{l c c c}
                \toprule
                \textbf{Method} & \textbf{5 sec} & \textbf{10 sec} & \textbf{20 sec} \\
                \midrule
                SVM        & 0.570 & 0.610 & 0.600 \\
                XGBoost    & 0.620 & 0.640 & 0.650 \\
                DNN        & 0.620 & 0.640 & 0.680 \\
                \midrule
                \multicolumn{4}{c}{\textbf{TCCT-Net Methods (no windowing needed)}} \\
                \midrule
                TCCT-Net (HP)  & \multicolumn{3}{c}{\underline{0.72}} \\
                TCCT-Net + VLM (HP) & \multicolumn{3}{c}{\textbf{0.74}} \\
                \bottomrule
            \end{tabular}
        }
        \caption{\textbf{PAFE}: Weighted F1 scores for different window sizes (SVM, XGBoost, DNN) and TCCT-Net methods (no windowing). Results for SVM, XGBoost, and DNN from \cite{Taeckyung2022PAFE}; TCCT-Net from \cite{Vedernikov_2024_CVPR}.}
        \label{tab:results_pafe}
    \end{subtable}
    \caption{Combined comparison of results across the \textit{EngageNet}, \textit{DREAMS}, and \textit{PAFE} datasets. Abbreviations: \textbf{EG} = Eye Gaze, \textbf{HP} = Head Pose, \textbf{AU} = Action Units, \textbf{LMK} = Landmarks, \textbf{PDM} = Point Distribution Model, \textbf{ST} = Single-Task, \textbf{MT} = Multi-Task, \textbf{TL} = Transfer Learning. \textbf{Bold} indicates the best performance and \underline{underline} the second-best.}
\end{table}

\section{Experiment Results}\label{sec:exp_results}
\subsection{Overall Performance}
\noindent \textbf{\textit{EngageNet dataset}}: Table~\ref{tab:results_engagenet} shows that conventional methods (LSTM, TCN, CNN-LSTM) achieve accuracies around $67\%–68\%$, while TCCT-Net reaches $68.91\%$ with Head Pose features. With VLM guidance, Head Pose accuracy slightly drops to $68.16\%$, whereas for Eye Gaze a modest improvement is observed ($64.61\%$ vs. $64.33\%$). In addition, the results with combinations of features further demonstrate the strength of VLM approach: when using Head Pose with Action Units, TCCT-Net with VLM outperforms its baseline ($66.29\%$ vs. $65.08\%$), and similarly, for the Action Units and Eye Gaze combination, VLM guidance brings the accuracy to $65.73\%$ compared to $65.17\%$. It suggests that while TCCT-Net is robust on its own, incorporating VLM guidance is very useful as it consistently leverages complementary information from different feature combinations to boost performance. To give more insight into class-wise performance, per-class confusion matrices for the vanilla TCCT-Net and our TCCT-Net + VLM are provided in Appendix~\ref{sec:conf_matrices}.

\vspace{0.05in} \noindent \textbf{\textit{DREAMS dataset}}: Table \ref{tab:results_dreams} reports weighted F$1$ scores for various behavioral feature combinations. Notably, TCCT-Net with VLM guidance outperforms both the vanilla TCCT-Net and baseline methods even with fewer features. Using only Eye Gaze features, TCCT-Net + VLM achieves a weighted F$1$ of $0.423$, and combining Eye Gaze with Head Pose reaches $0.434$. These results show that integrating VLM guidance improves performance and allows superior results with a reduced feature set, highlighting the effectiveness of our noise-robust training strategy. It is worth noting that TCCT-Net achieves a score of $0.185$ when using Head Pose alone or in combinations (with Action Units, Eye Gaze or both), indicating that the Head Pose feature by itself does not contribute to an improvement in performance. Moreover, the experiments reveal that fusing additional features does not always enhance the weighted F$1$ score, regardless of the method employed, which emphasizes the challenge of identifying complementary feature combinations.

\vspace{0.05in} \noindent \textbf{\textit{PAFE dataset}}: Traditional models (SVM, XGBoost, DNN) yield weighted F$1$ scores between $0.57$ and $0.68$, depending on the window size. In contrast, TCCT-Net with Head Pose features reaches $0.72$ (Table~\ref{tab:results_pafe}), and with VLM-guided strategy, it improves to $0.74$, demonstrating the power of such approaches in addressing subjective labeling. Importantly, this improvement comes without any hand-crafted fixed-length time windows, simplifying the pipeline.

\vspace{0.05in}
\noindent Across three datasets, the enhanced TCCT-Net, integrating curriculum learning, label refinement, weighted augmentation, and VLM guidance, consistently improves robustness and generalization. Although performance gains vary by dataset, our approach mitigates noisy, subjective labels, highlighting its potential for engagement recognition. In absolute terms, the improvements are modest (maximum of $+1.21\%$ on \textit{EngageNet} accuracy and $+0.22 / +0.06$ weighted F$1$ on \textit{DREAMS} / \textit{PAFE}), but they are obtained repeatedly across three independently collected datasets and several feature sets. With the exception of a single configuration (Head Pose alone on \textit{EngageNet}, $-0.75\%$), we do not observe notable drops, suggesting that the VLM-guided curriculum adds robustness to label subjectivity without destabilizing a strong baseline. These results therefore support the use of our framework as a reliability- and robustness-oriented enhancement rather than a mechanism for large accuracy jumps on any single benchmark.

\subsection{Ablation Study}
Table~\ref{tab:ablation} shows the impact of VLM-guided, curriculum learning–based training on TCCT-Net across multiple datasets. On the \textit{EngageNet} dataset, using only Eye Gaze features gives a baseline accuracy of $64.33\%$. With a one-stage VLM-guided approach, accuracy drops to $59.29\%$, but the two-stage curriculum raises it to $64.61\%$. Similarly, for the combination of Head Pose and Action Units, performance improves from $65.08\%$ to $66.29\%$, and for Eye Gaze with Action Units from $65.17\%$ to $65.73\%$. In the \textit{DREAMS} dataset, the weighted F$1$ score for Eye Gaze goes from $0.357$ to $0.423$, and for Head Pose it jumps from $0.185$ to $0.405$. Combining features, such as Eye Gaze with Head Pose, increases the score from $0.185$ to $0.434$, with similar improvements seen for the other combinations. On the \textit{PAFE} dataset, using Head Pose features, the weighted F$1$ score improves from a baseline of $0.72$ to $0.74$ with the two-stage method, compared to $0.65$ for the one-stage approach. Overall, these results show that our two-stage VLM-guided curriculum not only recovers but also improves the baseline performance, enhancing the model's ability to handle ambiguous and noisy labels and leading to a more robust and effective system across all datasets.

\begin{table}[h]
    \centering
    \resizebox{\textwidth}{!}{%
        \begin{tabular}{lccc}
            \toprule
            \textbf{Feature} & \textbf{TCCT-Net} & \makecell{\textbf{TCCT-Net + VLM}\\(CL: one stage)} & \makecell{\textbf{TCCT-Net + VLM}\\(CL: two stage)} \\
            \midrule
            \multicolumn{4}{c}{\textbf{\textit{EngageNet} (Accuracy \%)}} \\
            \midrule
            \textbf{EG}     & \underline{64.33} & 59.29 \textcolor{red}{\fontsize{5}{6}\selectfont$-5.04$} & \textbf{64.61} \textcolor{apricot}{\fontsize{5}{6}\selectfont$+0.28$} \\
            \textbf{HP + AU}  & \underline{65.08} & 61.62 \textcolor{red}{\fontsize{5}{6}\selectfont$-3.46$} & \textbf{66.29} \textcolor{apricot}{\fontsize{5}{6}\selectfont$+1.21$} \\
            \textbf{EG + AU}  & \underline{65.17} & 61.44 \textcolor{red}{\fontsize{5}{6}\selectfont$-3.73$} & \textbf{65.73} \textcolor{apricot}{\fontsize{5}{6}\selectfont$+0.56$} \\
            \midrule
            \multicolumn{4}{c}{\textbf{\textit{DREAMS} (Weighted F1)}} \\
            \midrule
            \textbf{EG}         & \underline{0.357} & 0.317 \textcolor{red}{\fontsize{5}{6}\selectfont$-0.040$} & \textbf{0.423} \textcolor{apricot}{\fontsize{5}{6}\selectfont$+0.066$} \\
            \textbf{HP}         & 0.185 & \underline{0.301} \textcolor{apricot}{\fontsize{5}{6}\selectfont$+0.116$} & \textbf{0.405} \textcolor{apricot}{\fontsize{5}{6}\selectfont$+0.220$} \\
            \textbf{AU}         & 0.316 & \underline{0.334} \textcolor{apricot}{\fontsize{5}{6}\selectfont$+0.018$} & \textbf{0.343} \textcolor{apricot}{\fontsize{5}{6}\selectfont$+0.027$} \\
            \textbf{EG + HP}      & 0.185 & \underline{0.288} \textcolor{apricot}{\fontsize{5}{6}\selectfont$+0.103$} & \textbf{0.434} \textcolor{apricot}{\fontsize{5}{6}\selectfont$+0.249$} \\
            \textbf{EG + AU}      & \underline{0.339} & 0.292 \textcolor{red}{\fontsize{5}{6}\selectfont$-0.047$} & \textbf{0.386} \textcolor{apricot}{\fontsize{5}{6}\selectfont$+0.047$} \\
            \textbf{HP + AU}      & 0.185 & \underline{0.284} \textcolor{apricot}{\fontsize{5}{6}\selectfont$+0.099$} & \textbf{0.316} \textcolor{apricot}{\fontsize{5}{6}\selectfont$+0.131$} \\
            \textbf{EG + HP + AU}   & 0.185 & \underline{0.354} \textcolor{apricot}{\fontsize{5}{6}\selectfont$+0.169$} & \textbf{0.384} \textcolor{apricot}{\fontsize{5}{6}\selectfont$+0.199$} \\
            \midrule
            \multicolumn{4}{c}{\textbf{\textit{PAFE} (Weighted F1)}} \\
            \midrule
            \textbf{HP}         & \underline{0.72}  & 0.65 \textcolor{red}{\fontsize{5}{6}\selectfont$-0.07$}  & \textbf{0.74} \textcolor{apricot}{\fontsize{5}{6}\selectfont$+0.02$} \\
            \bottomrule
        \end{tabular}}
    \caption{Ablation study comparing baseline TCCT-Net \cite{Vedernikov_2024_CVPR} with our curriculum learning-based, VLM-guided training, showing incremental performance gains on \textit{EngageNet}, \textit{DREAMS}, and \textit{PAFE} datasets. Abbreviations: \textbf{EG} = Eye Gaze, \textbf{HP} = Head Pose, \textbf{AU} = Action Units, \textbf{CL} = Curriculum Learning. \textbf{Bold} indicates the best performance and \underline{underline} the second-best.}
    \label{tab:ablation}
\end{table}

\section{Conclusion}\label{sec:conclusion}
This study has addressed the challenge of subjective annotations in engagement recognition by integrating VLM guidance into both dataset refinement and training strategies. By employing a structured questionnaire, our method effectively distinguishes high- from low-reliability samples, thereby enabling the selective training of classical models on cleaner data. Furthermore, introduction of a curriculum learning framework paired with soft label refinement allows models to progressively adapt to ambiguous cases, reducing overfitting to annotator-specific biases. Experimental results across multiple datasets demonstrate that our approach not only enhances overall classification accuracy but also improves robustness against label noise. Our ablation studies reveal that careful design of prompt detail and temporal context is crucial for maximizing VLM performance and downstream model benefits. Moving forward, we envision further integration of interactive VLM-assisted annotation protocols and development of evaluation metrics that better capture the inherent subjectivity of engagement. We will also validate our framework on the gaming benchmarks (including \textit{AGAIN} \cite{Melhart_2022} and others) to assess its generalizability in dynamic, interactive environments. Ultimately, our work sets the stage for more reliable affective computing systems by advancing data quality and noise-robust learning in subjective emotion recognition tasks.


\section*{Declarations}
\noindent \textbf{Acknowledgments}: The authors would like to acknowledge the support of the Center for Machine Vision and Signal Analysis, University of Oulu. \vspace{.2in}

\noindent \textbf{Funding}: This research received no external funding. \vspace{.2in}

\noindent \textbf{Competing interests}: The authors have no competing interests to declare. \vspace{.2in}

\noindent \textbf{Availability of data and material}: The datasets used in this study were obtained from the original authors upon request and are not publicly available due to data sharing restrictions. The authors do not have permission to redistribute these datasets. Data can be requested directly from the respective dataset owners.\vspace{.2in}

\noindent \textbf{Authors' contributions}: 
\textit{Alexander Vedernikov}: Conceptualization, Data Curation, Methodology, Software, Investigation, Formal Analysis, Visualization, Writing – Original Draft, Writing – Review \& Editing. 
\textit{Puneet Kumar}: Investigation, Formal Analysis, Writing – Review \& Editing. 
\textit{Haoyu Chen}: Investigation, Formal Analysis, Supervision, Writing – Review \& Editing. 
\textit{Tapio Seppänen}: Supervision, Writing – Review \& Editing. 
\textit{Xiaobai Li}: Supervision, Writing – Review \& Editing.
\vspace{.2in}

\noindent \textbf{Ethics approval}: Not applicable. \vspace{.2in}

\noindent \textbf{Consent for publication}: Not applicable.\vspace{.2in}


\begin{appendices}
\section{Computational Settings}\label{sec:Computational_Settings}
Computational tasks for VLM-based engagement classification (Section~\ref{sec:VLM_analysis}) were conducted on a supercomputer, utilizing between one and four Nvidia V$100$ GPUs, depending on model size. Classical computer vision methods (Section~\ref{sec:curriculum}) were processed on a supercomputer, utilizing $32$ SMT-enabled CPU cores and $32$ GB RAM, complemented by an AMD MI$250$ GPU with $128$GB of memory. We ensured consistency of the experiments by utilizing the identical computational environment.

\section{Conversation Mode Prompt Details}\label{sec:Conversation_Mode_Prompt}
In our few-shot/image Conversation mode experiments, InternVL$2$ was provided with a sequence of prompts to guide its analysis and classification of engagement levels. The prompting protocol consisted of the following stages:

\noindent \textbf{1. Example provision stage:} model was provided with a few example images for each engagement class, along with corresponding affective descriptions. These examples served to establish a reference for the further analysis. 

\noindent \textbf{2. Analysis stage:} next, model received a prompt instructing it to analyze a new image based on the previously provided examples: \textit{``Using the previous examples as a reference, analyze the person's affective state in the new image."}

\noindent \textbf{3. Classification stage:} finally, the model was prompted to output a numeric engagement level, where mapping is defined as: \textit{``Based on your analysis and observations, classify the engagement level in the new image. Output only the engagement level as a single number: $0$ - Not engaged, $1$ - Low engagement, $2$ - Moderate engagement, $3$ - High engagement. Provide no additional text or explanation in the output."}

\section{Full Questionnaire Details}\label{sec:Questionnaire_Details}
Table~\ref{tab:engagement_questions} presents the complete set of YES/NO questions used to guide the VLM analysis for engagement assessment. The questions are organized into four key categories: ($1$) Eye Gaze and Attention, ($2$) Facial Expressions, ($3$) Body Language and Posture, and ($4$) Distractions and External Activities. These targeted prompts were designed to capture specific, observable behaviors in order to reduce ambiguity and inter-annotator variability.

\begin{table}[!tbp]
  \centering
  {\fontsize{7.5}{9}\selectfont
    \setlength\tabcolsep{1pt}
  \begin{tabular}{p{1\textwidth}}
    \toprule
    \textbf{Specific Area and Questions} \\
    \midrule
    \textbf{Category \textbf{$\mathbf{1}$}: Eye Gaze and Attention} \\
    \textbf{Q$\mathbf{1}$:} Does the student frequently shift their gaze significantly away from the screen or primary point of focus? \\
    \textbf{Q$\mathbf{2}$:} Does the student’s eye gaze remain concentrated in one consistent direction while watching the video? \\
    \textbf{Q$\mathbf{3}$:} Does the student keep their attention directed at a single point during most of the video? \\
    \textbf{Q$\mathbf{4}$:} Does the student’s eye movement stay within a limited area, suggesting focused attention on the content? \\
    \textbf{Q$\mathbf{5}$:} Does the student frequently glance away from the screen? \\
    \midrule
    \textbf{Category \textbf{$\mathbf{2}$}: Facial Expressions} \\
    \textbf{Q$\mathbf{6}$:} Does the student appear uninterested or bored based on their facial expressions? \\
    \textbf{Q$\mathbf{7}$:} Does the student show signs of fatigue or sleepiness, such as yawning or nodding off? \\
    \textbf{Q$\mathbf{8}$:} Does the student barely open their eyes, appearing tired? \\
    \textbf{Q$9$:} Does the student appear to like or enjoy the content being presented? \\
    \midrule
    \textbf{Category \textbf{$\mathbf{3}$}: Body Language and Posture} \\
    \textbf{Q$\mathbf{10}$:} Is the student fidgeting restlessly in their chair? \\
    \textbf{Q$\mathbf{11}$:} Does the student exhibit a passive posture with minimal movement? \\
    \textbf{Q$\mathbf{12}$:} Does the student lean forward, appearing highly engaged with the content? \\
    \midrule
    \textbf{Category \textbf{$\mathbf{4}$}: Distractions and External Activities} \\
    \textbf{Q$\mathbf{13}$:} Is the student using a phone or other device during the lecture? \\
    \textbf{Q$\mathbf{14}$:} Does the student talk to others or engage in activities unrelated to the lecture? \\
    \textbf{Q$\mathbf{15}$:} Are there any signs of the student being distracted from the lecture? \\
    \bottomrule
  \end{tabular}
   \caption{Categorized areas and their associated questions for engagement assessment.}
  \label{tab:engagement_questions}
  }
\end{table}

\section{VLM-based Engagement Classification}\label{sec:vlm_eng_class}
Table~\ref{tab:starting_comparison_acc} summarizes the zero- and few-shot performance of VLMs on the \textit{EngageNet} validation set in both Single Question and Conversation modes. In Single-Question mode, the LLaVA model achieved the highest accuracy of $40.06\%$ with two images, while InternVL$2$ reached a peak of $28.66\%$. InternVL$2$’s conversation mode (absent in LLaVA) shows a modest improvement, with accuracy increasing from $23.44\%$ to $27.00\%$ in few-shot, single-image settings. This suggests that a single frame does not suffice for accurate engagement detection. Sequential questioning, by utilizing temporal context, shifts the analysis from isolated snapshots to a more dynamic and integrated evaluation of engagement. We then evaluated InternVL$2$ in conversation mode using video input, comparing the $8$B and $26$B models. The $8$B model achieved $35.20\%$ accuracy, while the $26$B model reached only $29.97\%$, suggesting that larger models may introduce additional noise. It further reached $45.28\%$ with structured question guidance. Thus, we adopt InternVL$2$ ($8$B) for future experiments. In summary, our experiments demonstrate that without clear, targeted prompts, VLMs can vary in accuracy by as much as $15\%$, highlighting the inherent challenges in engagement estimation due to both model limitations and human label ambiguity. These results emphasize the critical role of prompt clarity in achieving robust performance in nuanced tasks.

\begin{table}[!t]
  \centering
  \resizebox{.6\textwidth}{!}{
    \setlength\tabcolsep{3pt}
  \begin{tabular}{lcc}
    \toprule
    \textbf{Model} & \textbf{Input} & \begin{tabular}{@{}c@{}}\textbf{Validation} \\ \textbf{Accuracy (\%)} \end{tabular} \\
    \midrule
    \multicolumn{3}{c}{\textbf{Single Question Mode}} \\
    \midrule
    InternVL$2$ ($8$B) & $0$-shot/$1$ image & $23.44$ \\
                   & $0$-shot/$2$ images & \textbf{$\mathbf{28.66}$} \\
                   & $0$-shot/$3$ images & $27.45$ \\
                   & $0$-shot/$4$ images & $24.65$ \\
                   & $0$-shot/$5$ images & $24.28$ \\
    \midrule
    LLaVA $1.6$ ($7$B) & $0$-shot/$1$ image & $33.05$ \\
                   & $0$-shot/$2$ images & \textbf{$\mathbf{40.06}$} \\
                   & $0$-shot/$3$ images & $35.48$ \\
                   & $0$-shot/$4$ images & $35.11$ \\
                   & $0$-shot/$5$ images & $31.84$ \\
    \midrule
    VILA $1.5$ ($8$B)  & $0$-shot/$1$ image & $19.89$ \\
                   & $0$-shot/$2$ images & \textbf{$\mathbf{23.72}$} \\
                   & $0$-shot/$3$ images & $21.85$ \\
                   & $0$-shot/$4$ images & $23.34$ \\
                   & $0$-shot/$5$ images & $21.94$ \\
    \midrule
    \multicolumn{3}{c}{\textbf{Conversation Mode}} \\
    \midrule
    InternVL$2$ ($8$B) & few-shot/$1$ image & $27.00$ \\
                   & $0$-shot/video & $35.20$ \\
                   & \makecell{structured question\\\ guidance} & $\underline{\textbf{$\mathbf{45.28}$}}$ \\
    InternVL$2$ ($26$B) & $0$-shot/video & $29.97$ \\
    LLaVA$1.6$ ($7$B) & X & X \\
    VILA$1.5$ ($8$B) & X & X \\
    \bottomrule
  \end{tabular}}
  \caption{Zero-shot and few-shot performance of VLMs on \textit{EngageNet} validation set: four-class classification accuracy using single-question and conversation modes. \textbf{Bold} denotes best result per model in each mode, while \underline{\textbf{bold and underline}} denotes the best result across all experiments. X - Conversation mode isn't supported.}
  \label{tab:starting_comparison_acc}
\end{table}

\section{Enhancing Data Quality with VLM Filtering}\label{sec:Enhancing_Data}
Although the structured questionnaire design from Appendix~\ref{sec:Questionnaire_Details} helps target observable engagement cues, further refinement is required to reduce subjective label inconsistencies. Therefore, we used VLMs to divide the Validation set of the used datasets into so-called Accepted and Rejected subsets, enabling an in-depth evaluation of classical computer vision models. This approach ($1$) assesses how label quality impacts classical computer vision model performance, particularly in tasks as subjective as engagement classification, ($2$) highlights the impact of label quality on model robustness, and ($3$) provides insights into dataset reliability.

We employed the InternVL$2$ ($8$B) model to analyze the \textit{EngageNet} and \textit{PAFE} datasets, then split each into Accepted and Rejected subsets. The process was structured around a two-round conversation approach using video inputs. Video input involved eight frames extracted evenly across the entire video for the \textit{EngageNet} dataset and $20$ frames for the \textit{PAFE} dataset. In the first round, the model responded to a comprehensive set of YES/NO questions (see Table~\ref{tab:engagement_questions}) designed to detect key engagement indicators, such as eye movement consistency, signs of distraction, body posture, and expressions of interest or disinterest. After gathering this information, the model proceeded to a second round where it assigned each sample an engagement level on a scale from $0$ to $3$ for \textit{EngageNet} dataset case, and $0$ to $1$ for \textit{PAFE} dataset case. Based on the model’s predictions, we divided the Validation set of each dataset into an Accepted subset, where model classifications matched the ground truth labels, and Rejected subset, where discrepancies occurred.

\section{Comparative Analysis of Classical Computer Vision Models}\label{sec:cv_classic}
In engagement recognition, classical computer vision models may overfit to annotator biases, particularly when faced with subjective and inconsistent labels. VLM-based filtering aims to assess these classical computer vision models on subsets with varying label clarity, providing an opportunity to explore the effects of annotation reliability on model accuracy and to reveal dependencies on high-consensus labels. This part of experiment explores whether certain portions of the dataset better align with classical computer vision model expectations, offering an indirect assessment of annotation reliability. By examining performance across these subsets and using classical computer vision models (TCCT-Net \cite{Vedernikov_2024_CVPR}, ResNet + TCN \cite{abedi2021improving}, EfficientNet + LSTM \cite{selim2022students}, EfficientNet + Bi-LSTM \cite{selim2022students}; each trained on the \textit{EngageNet} and \textit{PAFE} datasets) we assess robustness across the entire validation set, as well as its Accepted and Rejected subsets. This method provides an approach, where VLM models help filter ambiguous labels and highlight areas for data quality improvement, systematically identifying and potentially filtering out unreliable annotations.

\section{Human Performance in Engagement Analysis}\label{sec:Human_Performance_Appendix}
To further examine the reliability of ground truth engagement labels, we conducted a human evaluation experiment, measuring annotator consistency with predefined labels and comparing it against VLM performance, highlighting areas of alignment and ambiguity. It provides a benchmark for assessing model performance on subjective tasks and highlights inherent challenges in engagement classification.

\vspace{.02in}\noindent \textbf{Participant selection:} We selected $10$ participants experienced in affective computing and familiar with emotion analysis, aiming to reduce variability and establish a realistic baseline for annotating subjective engagement data. Each participant received a sweet reward for participation.

\vspace{.02in}\noindent \textbf{Engagement classification task:} Participants were presented with $50$ video samples from the Validation set of the \textit{EngageNet} dataset and tasked with categorizing each one into predefined engagement levels ($0$ to $3$). To improve consistency in this subjective task, annotators followed detailed guidelines provided by the dataset authors. Each video sample could be viewed up to two times, with the goal of assigning engagement levels that matched or closely approximated the ground truth labels.

\vspace{.02in}\noindent \textbf{Tolerance-based accuracy metric:} To account for slight discrepancies in engagement levels, we introduced a `tolerance-based accuracy' metric, allowing a margin of error of ±$1$ level in classification, in addition to the standard \textit{EngageNet} accuracy metric. This approach provided insight into how closely human annotators could approximate the correct label without an exact match, offering a realistic perspective on performance in this subjective task.

\section{Extended Ablation Studies on the Validation Set}\label{sec:Ablation_Test}
To investigate the impact of subjective annotations and potential model biases, we conducted an ablation study analyzing both InternVL$2$ ($8$B) using a video-based zero-shot approach and the classical computer vision model TCCT-Net. It examined how variations in input configurations (specifically, the number of frames per video and the selection of specific questions from our structured guidance list (see Table~\ref{tab:engagement_questions}) affect the performance of both models in engagement classification. By assessing these factors, we aimed to identify optimal configurations that could mitigate the influence of subjective labels, thereby enhancing model robustness and improving classification accuracy for both VLMs and classical computer vision approaches in the context of subjective emotion recognition tasks. \vspace{0.1cm}

\begin{table}[!h]
  \centering
  {\fontsize{8.1}{9.5}\selectfont
  \setlength\tabcolsep{4pt}
  \begin{tabular}{@{}c c c c c c@{}}
    \toprule
    \raisebox{1.5ex}{\shortstack{\textbf{Frames /}\\\textbf{Questions}}} & \raisebox{1.5ex}{\shortstack{\textbf{Full Validation}\\\textbf{Accuracy [\%]}}} & \multicolumn{2}{c}{\shortstack{\textbf{Rejected}\\\textbf{Accuracy [\%] / }\\\textbf{Samples}}} & \multicolumn{2}{c}{\shortstack{\textbf{Accepted}\\\textbf{Accuracy [\%] /}\\\textbf{Samples}}} \\
    \midrule
    \textbf{F$\mathbf{2}$}  & $43.04$ & $62.13$ \textcolor{red}{\fontsize{5}{6}\selectfont$-6.78$} & $610$ & \textbf{$\mathbf{77.87}$} \textcolor{apricot}{\fontsize{5}{6}\selectfont$+8.96$} & $461$ \\
    \textbf{F$\mathbf{4}$}  & $42.11$ & $61.45$ \textcolor{red}{\fontsize{5}{6}\selectfont$-7.46$} & $620$ & \textbf{$\mathbf{79.16}$} \textcolor{apricot}{\fontsize{5}{6}\selectfont$+10.25$} & $451$ \\
    \textbf{F$\mathbf{8}$}  & $45.28$ & $58.02$ \textcolor{red}{\fontsize{5}{6}\selectfont$-10.89$} & $586$ & \textbf{$\mathbf{82.06}$} \textcolor{apricot}{\fontsize{5}{6}\selectfont$+13.15$} & $485$ \\
    \textbf{F$\mathbf{10}$} & $44.44$ & $56.13$ \textcolor{red}{\fontsize{5}{6}\selectfont$-12.78$} & $595$ & \textbf{$\mathbf{84.87}$} \textcolor{apricot}{\fontsize{5}{6}\selectfont$+15.96$} & $476$ \\
    \textbf{F$\mathbf{16}$} & $50.70$ & $50.19$ \textcolor{red}{\fontsize{5}{6}\selectfont$-18.72$} & $528$ & \textbf{$\mathbf{87.11}$} \textcolor{apricot}{\fontsize{5}{6}\selectfont$+18.20$} & $543$ \\
    \textbf{F$\mathbf{20}$} & $47.99$ & $52.78$ \textcolor{red}{\fontsize{5}{6}\selectfont$-16.13$} & $557$ & \textbf{$\mathbf{86.38}$} \textcolor{apricot}{\fontsize{5}{6}\selectfont$+17.47$} & $514$ \\
    \midrule
    \textbf{Q$\mathbf{3}$}  & $36.23$ & \textbf{$\mathbf{69.84}$} \textcolor{apricot}{\fontsize{5}{6}\selectfont$+0.93$} & $683$ & $67.27$ \textcolor{red}{\fontsize{5}{6}\selectfont$-1.64$} & $388$ \\
    \textbf{Q$\mathbf{6}$}  & $33.99$ & \textbf{$\mathbf{71.57}$} \textcolor{apricot}{\fontsize{5}{6}\selectfont$+2.66$} & $707$ & $63.74$ \textcolor{red}{\fontsize{5}{6}\selectfont$-5.17$} & $364$ \\
    \textbf{Q$\mathbf{9}$}  & $43.79$ & $55.48$ \textcolor{red}{\fontsize{5}{6}\selectfont$-13.43$} & $602$ & \textbf{$\mathbf{86.14}$} \textcolor{apricot}{\fontsize{5}{6}\selectfont$+17.23$} & $469$ \\
    \textbf{Q$\mathbf{12}$} & $45.56$ & $57.63$ \textcolor{red}{\fontsize{5}{6}\selectfont$-11.28$} & $583$ & \textbf{$\mathbf{82.38}$} \textcolor{apricot}{\fontsize{5}{6}\selectfont$+13.47$} & $488$ \\
    \textbf{Q$\mathbf{15}$} & $50.70$ & $50.19$ \textcolor{red}{\fontsize{5}{6}\selectfont$-18.72$} & $528$ & \textbf{$\mathbf{87.11}$} \textcolor{apricot}{\fontsize{5}{6}\selectfont$+18.20$} & $543$ \\
    \bottomrule
  \end{tabular}
  \caption{Performance of VLM over different frame counts (F) and prompt question (Q) counts. Red/green values indicate accuracy differences from the TCCT-Net \cite{Vedernikov_2024_CVPR} baseline of $68.91\%$. Total number of samples in the Validation set of \textit{EngageNet} is $1071$. \textbf{Bold} indicates the best performance.}
  \label{tab:ablation_merged}
  }
\end{table}

\noindent \textbf{Number of input frames.} The results, presented in Table \ref{tab:ablation_merged}, reveal that increasing the number of frames enhances the performance of both the VLM and TCCT-Net models. For the VLM, accuracy increases slightly with more frames, peaking at $50.70$\% with $16$ frames before declining at $20$ frames ($47.99$\%). This indicates that additional temporal information from more frames allows the VLM to capture dynamic engagement cues more effectively, up to a point where adding more frames may introduce redundancy or noise. The slight decline beyond $16$ frames suggests a saturation point where the model's capacity to utilize additional information is exceeded. Similarly, the accuracy on Accepted subset for TCCT-Net improves as the frame count increases, reaching $87.11$\% with $16$ frames. This improvement underscores the importance of temporal dynamics in engagement recognition, as more frames provide a richer temporal context for the model to learn from. However, for Rejected subset, TCCT-Net's accuracy decreases with more frames, suggesting that increased frame counts may amplify the impact of label noise on these samples. The reduction in the number of Rejected samples as frames increase (from $610$ with $2$ frames to $528$ with $16$ frames) supports the idea that increased temporal data helps reduce ambiguity in engagement classification. Overall, these findings indicate that there is an optimal number of frames (around $16$ in this case) that balances the benefits of additional temporal information with the potential drawbacks of redundancy and increased computational load.
\vspace{0.1cm}

\begin{table}[!b]
  \centering
  {\fontsize{9}{9.5}\selectfont
    \setlength\tabcolsep{3pt}
  \begin{tabular}{cl}
    \toprule
    \textbf{Number of Questions} & \textbf{Questions (by Number) Included} \\
    \midrule
    \textbf{$\mathbf{15}$} & All questions ($1$ to $15$) \\
    \textbf{$\mathbf{12}$} & $1$, $2$, $6$, $7$, $8$, $9$, $10$, $11$, $12$, $13$, $14$, $15$ \\
    \textbf{$\mathbf{9}$} & $1$, $2$, $6$, $7$, $9$, $10$, $12$, $13$, $14$ \\
    \textbf{$\mathbf{6}$} & $1$, $2$, $9$, $10$, $12$, $13$ \\
    \textbf{$\mathbf{3}$} & $1$, $12$, $13$ \\
    \bottomrule
  \end{tabular}
  \caption{Ablation study question configurations: subsets of questions selected for each experiment.}
  \label{tab:filtering_questions}}
\end{table}

\noindent \textbf{Number of questions in the prompt.} We show process of forming a logical approach to selecting which questions to include when reducing list from $15$ questions to fewer numbers for ablation study (Table \ref{tab:filtering_questions}). It ensures that the selection process is systematic and grounded in the relevance of each question to engagement prediction. To systematically reduce the number of questions, we: ($1$) Group the questions into categories based on the aspects of engagement they assess. ($2$) Identify possible redundancy among questions. ($3$) Rank the questions based on their importance and uniqueness. ($4$) Select questions that cover the broadest range of engagement indicators with minimal overlap. The results of ablation study are presented in Table \ref{tab:ablation_merged}. The analysis reveals that question count significantly impacts the performance of both VLM and TCCT-Net. For VLM, accuracy steadily increases from $36.23\%$ with $3$ questions to $50.70\%$ with the full $15$ questions, indicating that a broader set of questions aids in capturing engagement cues. Notably, TCCT-Net achieves $87.11\%$ accuracy on accepted samples with all $15$ questions, while accuracy drops as question count reduces, reaching $67.27\%$ with only $3$ questions. This indicates that a more detailed and diverse prompt enhances the effectiveness of the VLM filtering, resulting in a higher-quality Accepted subset that improves the downstream performance of TCCT-Net. The number of Rejected samples also decreases as the question count increases, highlighting the role of comprehensive prompts in identifying and filtering out samples with inconsistent or noisy annotations. These findings emphasize the importance of carefully designing prompts to include sufficient detail for capturing the multifaceted nature of engagement, thereby improving model robustness and classification accuracy.

\section{Per-Class Confusion Matrices}\label{sec:conf_matrices}
\begin{figure}[!h]
  \centering
   \includegraphics[width=1\linewidth]{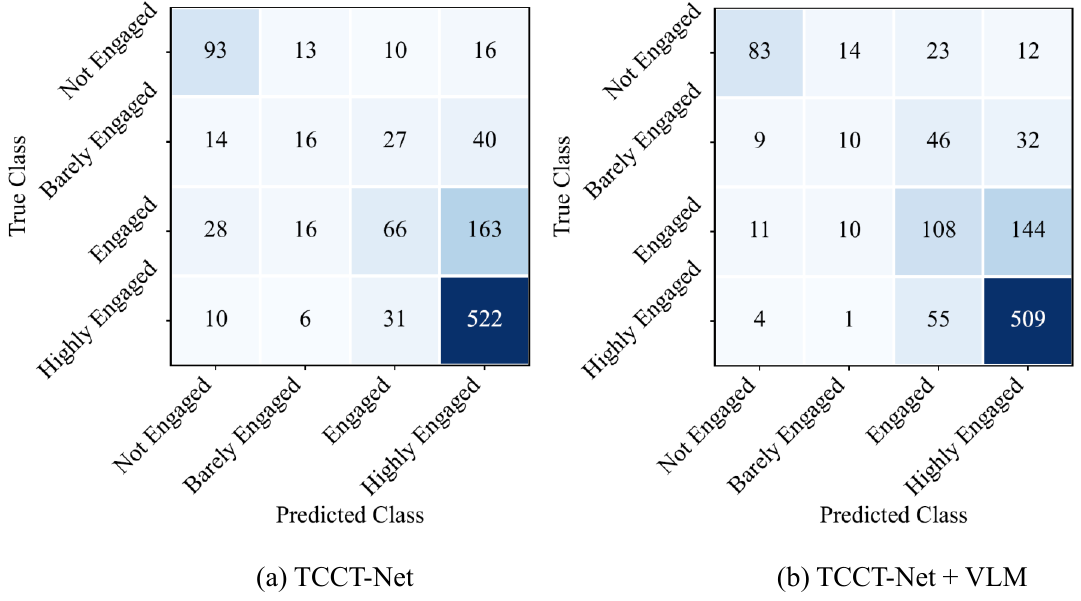} 
   \caption{Confusion matrices on the \textit{EngageNet} validation set using Action Units + Head Pose features. (a) Baseline TCCT-Net; (b) TCCT-Net augmented with VLM-guided curriculum learning and soft-label refinement.}
   \label{fig:conf_matrix_1} 
\end{figure}

\begin{figure}[!h]
  \centering
   \includegraphics[width=1\linewidth]{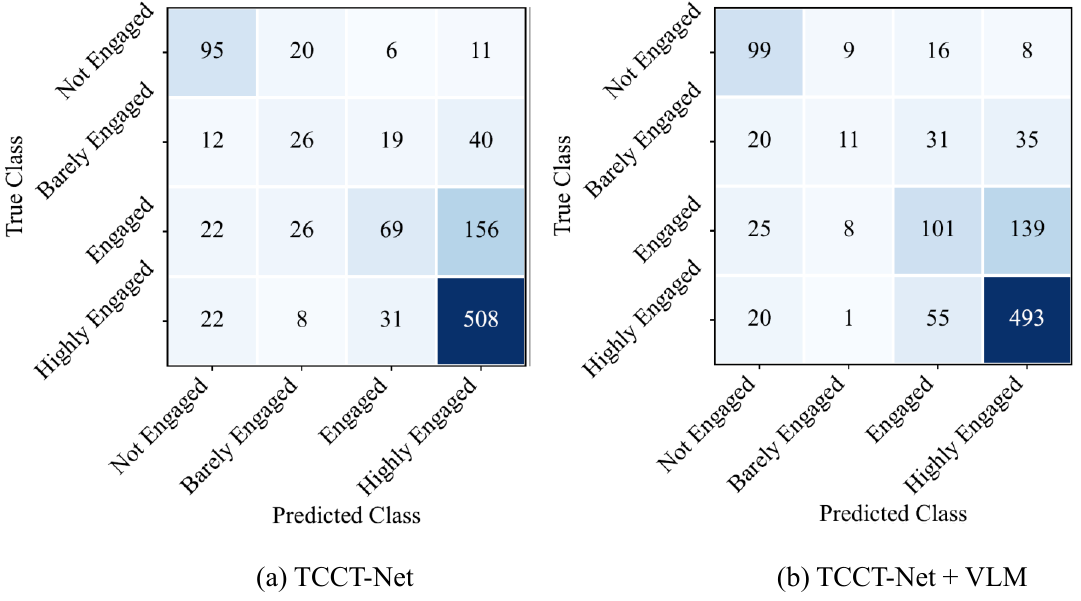}
   \caption{Confusion matrices on the \textit{EngageNet} validation set using Action Units + Eye Gaze features. (a) Baseline TCCT-Net; (b) TCCT-Net augmented with VLM-guided curriculum learning and soft‐label refinement.}
   \label{fig:conf_matrix_2} 
\end{figure}

Across both feature‐set configurations, integrating VLM guidance improves class discrimination, increasing true positives (diagonal entries) and cutting adjacent‐class confusions. In the Action Units + Head Pose case (Fig. \ref{fig:conf_matrix_1}), class $2$ (‘Engaged’) true positives rise from $66$ to $108$ ($+64\%$), while misclassifications into classes $1$ (‘Barely Engaged’) and $3$ (‘Highly Engaged’) each fall by $17$ and $19$ points, respectively, indicating clearer mid‐level engagement boundaries. Similarly, in the Action Units + Eye Gaze case (Fig. \ref{fig:conf_matrix_2}), class $2$ (‘Engaged’) true positives jump from $69$ to $101$ ($+46\%$), and class $3$ (‘Highly Engaged’) from $156$ to $139$ ($-17$ false positives), showing both stronger recall and reduced misclassifications into neighboring labels. Overall off-diagonal errors decrease by roughly $10–15\%$ across most classes, underscoring that VLM-filtered training both tightens core predictions and mitigates ambiguous label noise. Complementary qualitative analysis of these confusion matrices and representative clips indicates that the remaining errors are dominated by borderline cases. Many segments in which participants quietly follow the video or read questions with stable gaze and minimal facial movement are annotated as “Highly Engaged”, yet the VLM-guided model tends to assign them to “Engaged”, as the visible cues of enthusiasm (frequent expressions, stronger nods, pronounced forward lean) are weak. We also observe subject-specific differences in expressiveness: for some individuals, highly engaged behaviour remains visually subtle and is systematically under-estimated, whereas more animated participants are more reliably recognised as “Highly Engaged”. Taken together, these observations suggest that VLM guidance sharpens mid-scale boundaries while leaving residual ambiguity almost exclusively between neighbouring labels.

\end{appendices}

\bibliography{sn-bibliography}

@String(CVPR= {IEEE Conference Comput. Vis. Pattern Recog.})

@String(ICCV= {International Conference Comput. Vis.})

@String(ICPR = {International Conference Pattern Recog.})

@String(ICASSP=	{ICASSP})

@String(ICLR = {International Conference Learn. Represent.})

@String(CVPR  = {CVPR})

@String(ICCV  = {ICCV})

@String(ICPR  = {ICPR})

@String(ICLR  = {ICLR})

@InProceedings{Kumar_2025_WACV,
    author    = {Kumar, Puneet and others},
    title     = {{Multimodal Interpretable Depression Analysis using Visual Physiological Audio and Textual Data}},
    booktitle = {WACV},
    year      = {2025}
}

@article{kumar2022vistanet,
  title={{VISTANet: VIsual Spoken Textual Additive Net for interpretable multimodal emotion recognition}},
  author={Kumar, Puneet and others},
  journal={arXiv:2208.11450},
  year={2024}
}

@misc{han2021survey,
      title={A Survey of Label-noise Representation Learning: Past, Present and Future}, 
      author={Bo Han and Quanming Yao and Tongliang Liu and Gang Niu and Ivor W. Tsang and James T. Kwok and Masashi Sugiyama},
      year={2021},
      eprint={2011.04406},
      archivePrefix={arXiv},
      primaryClass={cs.LG},
      url={https://arxiv.org/abs/2011.04406}, 
}

@misc{song2022learn,
      title={Learning from Noisy Labels with Deep Neural Networks: A Survey}, 
      author={Hwanjun Song and Minseok Kim and Dongmin Park and Yooju Shin and Jae-Gil Lee},
      year={2022},
      eprint={2007.08199},
      archivePrefix={arXiv},
      primaryClass={cs.LG},
      url={https://arxiv.org/abs/2007.08199}, 
}

@inproceedings{arazo2019unsupervised,
  title={Unsupervised label noise modeling and loss correction},
  author={Arazo, Eric and Ortego, Diego and Albert, Paul and O’Connor, Noel and McGuinness, Kevin},
  booktitle={International conference on machine learning},
  pages={312--321},
  year={2019},
  organization={PMLR}
}

@article{Melhart_2022,
   title={The Arousal Video Game AnnotatIoN (AGAIN) Dataset},
   volume={13},
   ISSN={2371-9850},
   url={http://dx.doi.org/10.1109/TAFFC.2022.3188851},
   DOI={10.1109/taffc.2022.3188851},
   number={4},
   journal={IEEE Transactions on Affective Computing},
   publisher={Institute of Electrical and Electronics Engineers (IEEE)},
   author={Melhart, David and Liapis, Antonios and Yannakakis, Georgios N.},
   year={2022},
   month=oct, pages={2171–2184} }

@InProceedings{Karim_2022_CVPR,
    author    = {Karim, Nazmul and others},
    title     = {{UniCon: Combating Label Noise Through Uniform Selection and Contrastive Learning}},
    booktitle = {CVPR},
    year      = {2022},
    pages     = {9676-9686}
}

@article{plank2022problemhumanlabelvariation,
      title={{The `Problem' of Human Label Variation: On Ground Truth in Data, Modeling and Evaluation}}, 
      author={Barbara Plank},
      year={2022},
      eprint={2211.02570},
    journal={arXiv:2211.02570}
}

@INPROCEEDINGS{chou2019every,
  author={Chou, Huang-Cheng and Lee, Chi-Chun},
  booktitle={ICASSP 2019 - 2019 IEEE International Conference on Acoustics, Speech and Signal Processing (ICASSP)}, 
  title={Every Rating Matters: Joint Learning of Subjective Labels and Individual Annotators for Speech Emotion Classification}, 
  year={2019},
  volume={},
  number={},
  pages={5886-5890},
  doi={10.1109/ICASSP.2019.8682170}}

@inproceedings{wu2024can,
  title        = {Can Modelling Inter-Rater Ambiguity Lead To Noise-Robust Continuous Emotion Predictions?},
  author       = {Wu, Ya-Tse and Chang, Ching-Feng and Lee, Min-Ju and Kumar, Anil and Ruan, Jing and Smith, Jennifer},
  booktitle    = {Proceedings of Interspeech 2024},
  year         = {2024}
}

@INPROCEEDINGS{wu2022novel,
  author={Wu, Jingyao and Dang, Ting and Sethu, Vidhyasaharan and Ambikairajah, Eliathamby},
  booktitle={ICASSP 2022 - 2022 IEEE International Conference on Acoustics, Speech and Signal Processing (ICASSP)}, 
  title={A Novel Sequential Monte Carlo Framework for Predicting Ambiguous Emotion States}, 
  year={2022},
  volume={},
  number={},
  pages={8567-8571},
  doi={10.1109/ICASSP43922.2022.9746350}}

@article{hu2024unified,
      title={{Towards Unified Facial Action Unit Recognition Framework by Large Language Models}}, 
      author={Guohong Hu and Xing Lan and others},
      year={2024},
      journal={arXiv:2409.08444}, 
}

@article{shvetsova2024how,
      title={{HowToCaption: Prompting LLMs to Transform Video Annotations at Scale}}, 
      author={Nina Shvetsova and Anna Kukleva and others},
      year={2024},
      journal={arXiv: 2310.04900}, 
}

@article{zhang2024AC,
      title={{Affective Computing in the Era of Large Language Models: A Survey from the NLP Perspective}}, 
      author={Yiqun Zhang and and others},
      year={2024},
      journal={arXiv:2408.04638}, 
}

@article{xie2024emov,
      title={{EmoVIT: Revolutionizing Emotion Insights with Visual Instruction Tuning}}, 
      author={Hongxia Xie and ChuJun Peng and others},
      year={2024},
      journal={arXiv:2404.16670}, 
}

@article{etesam2024cont,
      title={{Contextual Emotion Recognition using Large Vision Language Models}}, 
      author={Yasaman Etesam and Özge Nilay Yalçın and others},
      year={2024},
      journal={arXiv:2405.08992}, 
}

@article{lu2024gptp,
      title={{GPT as Psychologist? Preliminary Evaluations for GPT-4V on Visual Affective Computing}}, 
      author={Hao Lu and Xuesong Niu and others},
      year={2024},
      journal={arXiv:2403.05916}, 
}

@ARTICLE{Nadeem20242566,
	author = {Nadeem, Mohammad and Sohail, Shahab Saquib and others},
	title = {{Vision-Enabled Large Language and Deep Learning Models for Image-Based Emotion Recognition}},
	year = {2024},
	journal = {Cognitive Computation},
	volume = {16},
	number = {5},
	pages = {2566 – 2579},
	type = {Article}
}

@article{yang2024emoll,
      title={{EmoLLM: Multimodal Emotional Understanding Meets Large Language Models}}, 
      author={Qu Yang and Mang Ye and Bo Du},
      year={2024},
      article={arXiv:2406.16442}, 
}

@article{mckinzie2024mm1,
      title={{MM1: Methods, Analysis \& Insights from Multimodal LLM Pre-training}}, 
      author={Brandon McKinzie and Zhe Gan and others},
      year={2024},
      eprint={2403.09611},
      archivePrefix={arXiv},
      primaryClass={cs.CV},
      journal={arXiv:2403.09611}, 
}

@article{northcutt2021pervasivelabelerrorstest,
      title={{Pervasive Label Errors in Test Sets Destabilize Machine Learning Benchmarks}}, 
      author={Curtis G. Northcutt and others},
      year={2021},
      eprint={2103.14749},
      archivePrefix={arXiv},
      primaryClass={stat.ML},
      journal={arXiv:2103.14749}, 
}

@inproceedings{
wen2023benign,
title={{Benign Overfitting in Classification: Provably Counter Label Noise with Larger Models}},
author={Kaiyue Wen and others},
booktitle={ICLR},
year={2023}
}

@InProceedings{Tu_2023_CVPR,
    author    = {Tu, Yuanpeng and others},
    title     = {Learning From Noisy Labels With Decoupled Meta Label Purifier},
    booktitle = {CVPR},
    year      = {2023},
    pages     = {19934-19943}
}

@article{khan2023novellossfunctionutilizing,
      title={{A Novel Loss Function Utilizing Wasserstein Distance to Reduce Subject-Dependent Noise for Generalizable Models in Affective Computing}}, 
      author={Nibraas Khan and others},
      year={2023},
      journal={arXiv:2308.10869}, 
}

@ARTICLE{Chen20222106,
	author = {Chen, Tao and Guo, Yanrong and Hao, Shijie and Hong, Richang},
	title = {{Exploring Self-Attention Graph Pooling With EEG-Based Topological Structure and Soft Label for Depression Detection}},
	year = {2022},
	journal = {IEEE Transactions on Affective Computing},
	volume = {13},
	number = {4},
	pages = {2106-2118}
}

@ARTICLE{Jing_10321660,
  author={Jiang, Jing and Deng, Weihong},
  journal={IEEE Transactions on Affective Computing}, 
  title={{Improving Multi-Label Facial Expression Recognition With Consistent and Distinct Attentions}}, 
  year={2024},
  volume={15},
  number={3},
  pages={1279-1288},
  doi={10.1109/TAFFC.2023.3333874}}

@ARTICLE{Chen_2024_static,
	author = {Chen, Yin and Li, Jia and Shan, Shiguang and Wang, Meng and others},
	title = {{From Static to Dynamic: Adapting Landmark-Aware Image Models for Facial Expression Recognition in Videos}},
	year = {2024},
	journal = {IEEE Transactions on Affective Computing},
	doi = {10.1109/TAFFC.2024.3453443}, 
	type = {Article},
	publication_stage = {Article in press},
	source = {Scopus}
}

@ARTICLE{Mao2022818,
	author = {Mao, Longbiao and Yan, Yan and Xue, Jing-Hao and Wang, Hanzi},
	title = {{Deep Multi-Task Multi-Label CNN for Effective Facial Attribute Classification}},
	year = {2022},
	journal = {IEEE Transactions on Affective Computing}, 
	doi = {10.1109/TAFFC.2020.2969189}, 
	type = {Article},
	publication_stage = {Final},
	source = {Scopus}
}

@article{zhang2023partiallabellearningemotion,
      title={{Partial Label Learning for Emotion Recognition from EEG}}, 
      author={Guangyi Zhang and Ali Etemad},
      year={2023},
      journal={arXiv:2302.13170}, 
}

@ARTICLE{Liu_10214365,
  author={Liu, Yang and others},
  journal={IEEE Transactions on Multimedia}, 
  title={{Uncertain Facial Expression Recognition via Multi-Task Assisted Correction}}, 
  year={2024},
  volume={26},
  number={},  
  doi={10.1109/TMM.2023.3301209}}

@inproceedings{Liu_2022,
   title={{Uncertain Label Correction via Auxiliary Action Unit Graphs for Facial Expression Recognition}}, 
   DOI={10.1109/icpr56361.2022.9956650},
   booktitle={ICPR},
   author={Liu, Yang and others},
   year={2022},
    pages={777–783} 
}

@ARTICLE{Jin20247395,
	author = {Jin, Zhigang and others},
	title = {{A Psychological Evaluation Method Incorporating Noisy Label Correction Mechanism}},
	year = {2024},
	journal = {Soft Computing},
	volume = {28},
	number = {11-12},
	doi = {10.1007/s00500-023-09479-w}, 
	type = {Article},
	publication_stage = {Final},
	source = {Scopus}
}

@Article{chen_13193891,
AUTHOR = {Chen, Tao and Zhang, Dong and Lee, Dah-Jye},
TITLE = {{A New Joint Training Method for Facial Expression Recognition with Inconsistently Annotated and Imbalanced Data}},
JOURNAL = {Electronics},
VOLUME = {13},
YEAR = {2024},
NUMBER = {19},
ARTICLE-NUMBER = {3891}
}

@ARTICLE{Tan2024,
	author = {Tan, Yumei and Xia, Haiying and Song, Shuxiang},
	title = {{Robust Consistency Learning for Facial Expression Recognition Under Label Noise}},
	year = {2024},
	journal = {Visual Computer},
	doi = {10.1007/s00371-024-03558-1}, 
	type = {Article},
	publication_stage = {Article in press},
	source = {Scopus}
}

@ARTICLE{Washington20211363,
	author = {Washington, Peter and Kalantarian, Haik and Kent, Jack and others},
	title = {{Training Affective Computer Vision Models by Crowdsourcing Soft-Target Labels}},
	year = {2021},
	journal = {Cognitive Computation},
	volume = {13},
	number = {5},
	pages = {1363-1373},
	doi = {10.1007/s12559-021-09936-4}, 
	type = {Article},
	publication_stage = {Final},
	source = {Scopus}
}

@INPROCEEDINGS{Darshan_9607483,
  author={Gera, Darshan and others},
  booktitle={ICCV Workshops}, 
  title={{Noisy Annotations Robust Consensual Collaborative Affect Exp. Recognition}}, 
  year={2021},
  volume={},
  number={},
  pages={3578-3585}
}

@ARTICLE{Zhou_PR_PL_10160130,
  author={Zhou, Rushuang and Zhang, Zhiguo and Fu, Hong and Zhang, Li and others},
  journal={IEEE Transactions on Affective Computing}, 
  title={{PR-PL: A Novel Prototypical Representation Based Pairwise Learning Framework for Emotion Recognition Using EEG}}, 
  year={2024},
  volume={15},
  number={2},
  pages={657-670}
}

@inproceedings{Khan2024111,
author = {Khan, Shehroz and Safa, Sadaf},
title = {{Revisiting Annotations in Online Student Engagement}},
year = {2024},
isbn = {9798400709319},
booktitle = {International Conference on Computing and Data Engineering},
pages = {111–117},
numpages = {7},
}

@ARTICLE{Zhong20221290,
	author = {Zhong, Peixiang and Wang, Di and Miao, Chunyan},
	title = {{EEG-Based Emotion Recognition Using Regularized Graph Neural Networks}},
	year = {2022},
	journal = {IEEE Transactions on Affective Computing},
	volume = {13},
	number = {3},
	pages = {1290-1301},
	doi = {10.1109/TAFFC.2020.2994159}, 
	type = {Article},
	publication_stage = {Final},
	source = {Scopus}
}

@ARTICLE{Jiang20232402,
	author = {Jiang, Jing and others},
	title = {{Boosting Facial Exp. Recognition by A Semi-Supervised Progressive Teacher}},
	year = {2023},
	journal = {IEEE Transactions on Affective Computing}, 
	doi = {10.1109/TAFFC.2021.3131621}, 
	type = {Article},
	publication_stage = {Final},
	source = {Scopus}
}

@ARTICLE{Liao20216609,
	author = {Liao, Jiacheng and others},
	title = {{Deep Facial Spatiotemporal Net For Engagement Prediction in Online Learning}},
	year = {2021},
	journal = {Applied Intelligence},
	volume = {51},
	number = {10},
	doi = {10.1007/s10489-020-02139-8}, 
	type = {Article},
	publication_stage = {Final},
	source = {Scopus}
}

@ARTICLE{Abedi20233535,
	author = {Abedi, Ali and others},
	title = {{Detecting Disengagement in Virtual Learning as an Anomaly using Temporal Convolutional Net. Autoencoder}},
	year = {2023},
	journal = {Signal, Image and Video Processing},
	volume = {17},
	number = {7},
	pages = {3535-3543},
	doi = {10.1007/s11760-023-02578-z}, 
	type = {Article},
	publication_stage = {Final},
	source = {Scopus}
}

@InProceedings{Vedernikov_2024_CVPR,
    author    = {Vedernikov, Alexander and others},
    title     = {{TCCT-Net: Two-Stream Network Architecture for Fast and Efficient Engagement Estimation via Behavioral Feature Signals}},
    booktitle = {CVPR Workshops},
    month     = {June},
    year      = {2024},
    pages     = {4723-4732}
}

@ARTICLE{Noroozi2021505,
	author = {Noroozi, Fatemeh and Corneanu, Ciprian Adrian and Kaminska, Dorota and Sapinski, Tomasz and Escalera, Sergio and others},
	title = {{Survey on Emotional Body Gesture Recognition}},
	year = {2021},
	journal = {IEEE Transactions on Affective Computing},
	volume = {12},
	number = {2},
	pages = {505 – 523}
}

@InProceedings{szegedy2016rethinking,
author = {Szegedy, Christian and others},
title = {{Rethinking the Inception Architecture for Computer Vision}},
booktitle = {CVPR},
year = {2016}
}

@article{zhou2012ensemble,
author = {Zhou, Zhi-Hua},
title = {Ensemble methods: Foundations and algorithms},
year = {2012},
journal = {CRC Press, Taylor \& Francis Group},
pages = {1–18},
doi = {10.1201/b12207}, 
type = {Book},
publication_stage = {Final},
source = {Scopus}
}

@inproceedings{bengio2009curriculum,
author = {Bengio, Yoshua and others},
title = {{Curriculum Learning}},
year = {2009},
isbn = {9781605585161},  
doi = {10.1145/1553374.1553380},
booktitle = {International Conference on Machine Learning (ICML)},
pages = {41–48},
numpages = {8}
}

@inproceedings{Singh2024dreams,
  author    = {Monisha Singh and Gulshan Sharma and others},
  title     = {{DREAMS: Diverse Reactions of Engagement and Attention Mind States Dataset}},
  booktitle = {ICPR},
  year      = {2024},
  pages     = {163--179}
}

@ARTICLE{selim2022students,
author = {Selim, Tasneem and others},
title = {{Students Engagement Level Detection in Online e-Learning Using Hybrid EfficientNetB7 Together With TCN, LSTM, and Bi-LSTM}},
year = {2022},
journal = {IEEE Access},
volume = {10},
pages = {99573-99583},
}

@CONFERENCE{abedi2021improving,
author = {Abedi, Ali and Khan, Shehroz S.},
title = {{Improving state-of-the-art in Detecting Student Engagement with Resnet and TCN Hybrid Network}},
year = {2021},
booktitle = { Conference on Robots and Vision},
pages = {151 – 157},
}

@inproceedings{dhall2023emotiw,
author = {Dhall, Abhinav and Singh, Monisha and Goecke, Roland and Gedeon, Tom and Zeng, Donghuo and others},
title = {{EmotiW 2023: Emotion Recognition in the Wild Challenge}},
year = {2023}, 
pages = {746–749},
numpages = {4}
}

@inproceedings{singh2023do,
author = {Singh, Monisha and others},
title = {{Do I Have Your Attention: A Large Scale Engagement Prediction Dataset and Baselines}},
year = {2023}, 
doi = {10.1145/3577190.3614164},
booktitle = {ICMI},
pages = {174–182},
numpages = {9}
}

@article{chen2023internvl,
      title={{InternVL: Scaling up Vision Foundation Models and Aligning for Generic Visual-Linguistic Tasks}},
      author={Chen, Zhe and others},
      journal={arXiv:2312.14238},
      year={2023}
  }

@article{chen2024far,
    title={{How Far Are We to GPT-4V? Closing the Gap to Commercial Multimodal Models with Open-Source Suites}},
    author={Chen, Zhe and Wang, Weiyun and Tian, Hao and Ye, Shenglong and others},
    journal={arXiv:2404.16821},
    year={2024}
  }

@misc{liu2024llavanext,
    title={{LLaVA-NeXT: Improved Reasoning, OCR And World Knowledge}},
    url={https://llava-vl.github.io/blog/2024-01-30-llava-next/},
    author={Liu, Haotian and Li, Chunyuan and Li, Yuheng and Li, Bo and Zhang, Yuanhan and others},
    month={January},
    year={2024}
}

@article{liu2023improvedllava,
      title={{Improved Baselines with Visual Instruction Tuning}}, 
      author={Liu, Haotian and Li, Chunyuan and Li, Yuheng and Lee, Yong Jae},
      journal={arXiv:2310.03744},
      year={2023},
}

@misc{liu2023llava,
      title={{Visual Instruction Tuning}}, 
      author={Liu, Haotian and Li, Chunyuan and Wu, Qingyang and Lee, Yong Jae},
      publisher={NeurIPS},
      year={2023},
}

@article{lin2023vila,
      title={{VILA: On Pre-training for Visual Language Models}},
      author={Ji Lin and Hongxu Yin and Wei Ping and Yao Lu and Pavlo Molchanov and others},
      year={2023},
      journal={arXiv:2312.07533}
}

@INPROCEEDINGS{Taeckyung2022PAFE,
  author={Lee, Taeckyung and Kim, Dain and Park, Sooyoung and others},
  booktitle={CVPR Workshops}, 
  title={{Predicting Mind-Wandering with Facial Videos in Online Lectures}}, 
  year={2022},
  volume={},
  number={},
  pages={2103-2112},
}

@misc{zhang2024adaneg,
      title={AdaNeg: Adaptive Negative Proxy Guided OOD Detection with Vision-Language Models}, 
      author={Yabin Zhang and Lei Zhang},
      year={2024},
      eprint={2410.20149},
      archivePrefix={arXiv},
      primaryClass={cs.CV},
      url={https://arxiv.org/abs/2410.20149}, 
}

@inproceedings{NEURIPS2024wang,
 author = {Wang, Haoyu and Huang, Zhuo and Lin, Zhiwei and Liu, Tongliang},
 booktitle = {Advances in Neural Information Processing Systems},
 doi = {10.52202/079017-3819},
 editor = {A. Globerson and L. Mackey and D. Belgrave and A. Fan and U. Paquet and J. Tomczak and C. Zhang},
 pages = {120159--120183},
 publisher = {Curran Associates, Inc.},
 title = {NoiseGPT: Label Noise Detection and Rectification through Probability Curvature},
 url = {https://proceedings.neurips.cc/paper\_files/paper/2024/file/d95cb79a3421e6d9b6c9a9008c4d07c5-Paper-Conference.pdf},
 volume = {37},
 year = {2024}
}

@misc{huang2024mmultimodal,
      title={Machine Vision Therapy: Multimodal Large Language Models Can Enhance Visual Robustness via Denoising In-Context Learning}, 
      author={Zhuo Huang and Chang Liu and Yinpeng Dong and Hang Su and Shibao Zheng and Tongliang Liu},
      year={2024},
      eprint={2312.02546},
      archivePrefix={arXiv},
      primaryClass={cs.CV},
      url={https://arxiv.org/abs/2312.02546}, 
}

@INPROCEEDINGS{Teotia10917661,
  author={Teotia, Jayant and Zhang, Xulang and Mao, Rui and Cambria, Erik},
  booktitle={2024 IEEE International Conference on Data Mining Workshops (ICDMW)}, 
  title={Evaluating Vision Language Models in Detecting Learning Engagement}, 
  year={2024},
  volume={},
  number={},
  pages={496-502},
  doi={10.1109/ICDMW65004.2024.00069}}

@misc{wang2025usingvisionlanguagemodels,
      title={Using Vision Language Models to Detect Students' Academic Emotion through Facial Expressions}, 
      author={Deliang Wang and Chao Yang and Gaowei Chen},
      year={2025},
      eprint={2506.10334},
      archivePrefix={arXiv},
      primaryClass={cs.CV},
      url={https://arxiv.org/abs/2506.10334}, 
}

@article{ekman1992argument,
  title={{A}n {A}rgument for {B}asic {E}motions},
  author={Ekman, Paul and others},
  journal={Cognition \& Emotion},
  volume={6},
  number={3-4},
  pages={169--200},
  year={1992},
  publisher={Taylor \& Francis}
}

@article{kumar2024nontypical,
  title={{Computational Analysis of Stress, Depression and Engagement in Mental Health}},
  author={Kumar, Puneet and others},
  journal={arXiv:2403.08824},
  year={2024}
}

\end{document}